\documentclass[10pt,twocolumn,letterpaper]{article}

\usepackage{cvpr}              

\usepackage[dvipsnames]{xcolor}

\definecolor{cvprblue}{rgb}{0.21,0.49,0.74}
\usepackage[pagebackref,breaklinks,colorlinks,citecolor=cvprblue]{hyperref}
\def\paperID{2479} 
\def\confName{CVPR}
\def\confYear{2025}
\usepackage{makecell,pifont,multirow,multicol}
\usepackage{colortbl}
\usepackage{caption}
\usepackage{subcaption}
\usepackage[misc]{ifsym}

\usepackage{hyperref}

\usepackage{orcidlink}

\def\ModelName{EgoFSD}

\def\logo{\makebox[30pt][l]{\raisebox{-1.2ex}{\includegraphics[height=28pt]{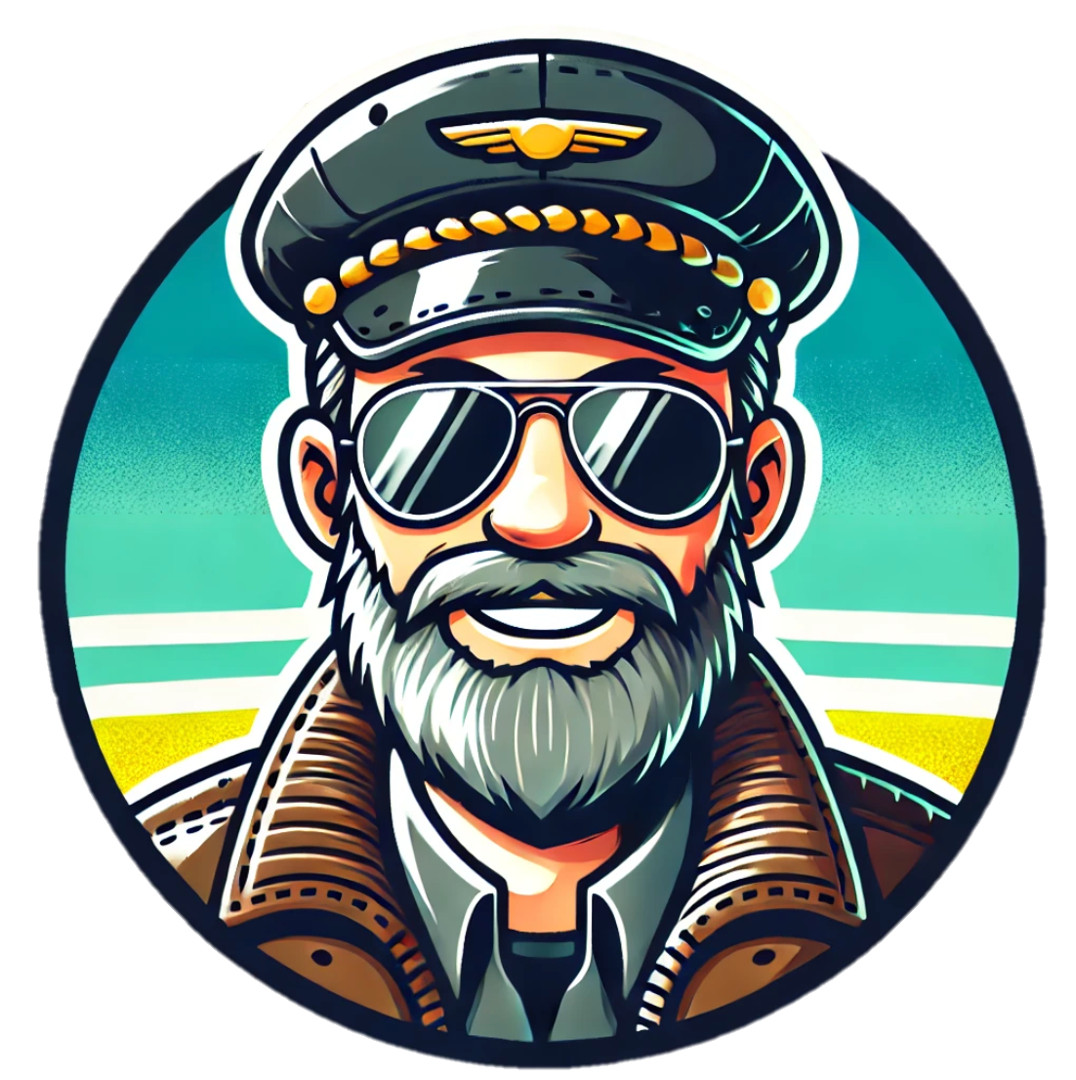}}}}

\title{\logo\ModelName: Ego-Centric \underline{F}ully \underline{S}parse Paradigm with Uncertainty Denoising and Iterative Refinement for Efficient End-to-End Self-\underline{D}riving}

\author{Haisheng Su$^{1}$ \quad
Wei Wu$^{2}$ \quad
Zhenjie Yang$^{1}$ \quad
Isabel Guan$^{3,}$$^{\textrm{\Letter}}$ \\
$^{1}$School of Computer Science, Shanghai Jiao Tong University \\ $^{2}${SenseAuto}, $^{3}${The Hong Kong University of Science and Technology} \\ 
{\tt\small suhaisheng@sjtu.edu.cn, eeguan@ust.hk} \\
\vspace{-0.4cm}
}

\begin{document}
\maketitle

\begin{abstract}

Current End-to-End Autonomous Driving (E2E-AD) methods resort to unifying modular designs for various tasks (e.g. perception, prediction and planning). Although optimized with a fully differentiable framework in a planning-oriented manner, existing end-to-end driving systems lacking ego-centric designs still suffer from unsatisfactory performance and inferior efficiency, due to rasterized scene representation learning and redundant information transmission. In this paper, we propose an ego-centric fully sparse paradigm, named EgoFSD, for end-to-end self-driving. Specifically, EgoFSD consists of sparse perception, hierarchical interaction and iterative motion planner. The sparse perception module performs detection and online mapping based on sparse representation of the driving scene. The hierarchical interaction module aims to select the Closest In-Path Vehicle / Stationary (CIPV / CIPS) from coarse to fine, benefiting from an additional geometric prior. As for the iterative motion planner, both selected interactive agents and ego-vehicle are considered for joint motion prediction, where the output multi-modal ego-trajectories are optimized in an iterative fashion. In addition, position-level motion diffusion and trajectory-level planning denoising are introduced for uncertainty modeling, thereby enhancing the training stability and convergence speed. Extensive experiments are conducted on nuScenes and Bench2Drive datasets, which significantly reduces the average L2 error by \textbf{59\%} and collision rate by \textbf{92\%} than UniAD while achieves \textbf{6.9$\times$} faster running efficiency.




\end{abstract}
    
\vspace{-0.3cm}
\section{Introduction}
\label{sec:intro}

\begin{figure}[t]
\centering
\includegraphics[width=8cm]{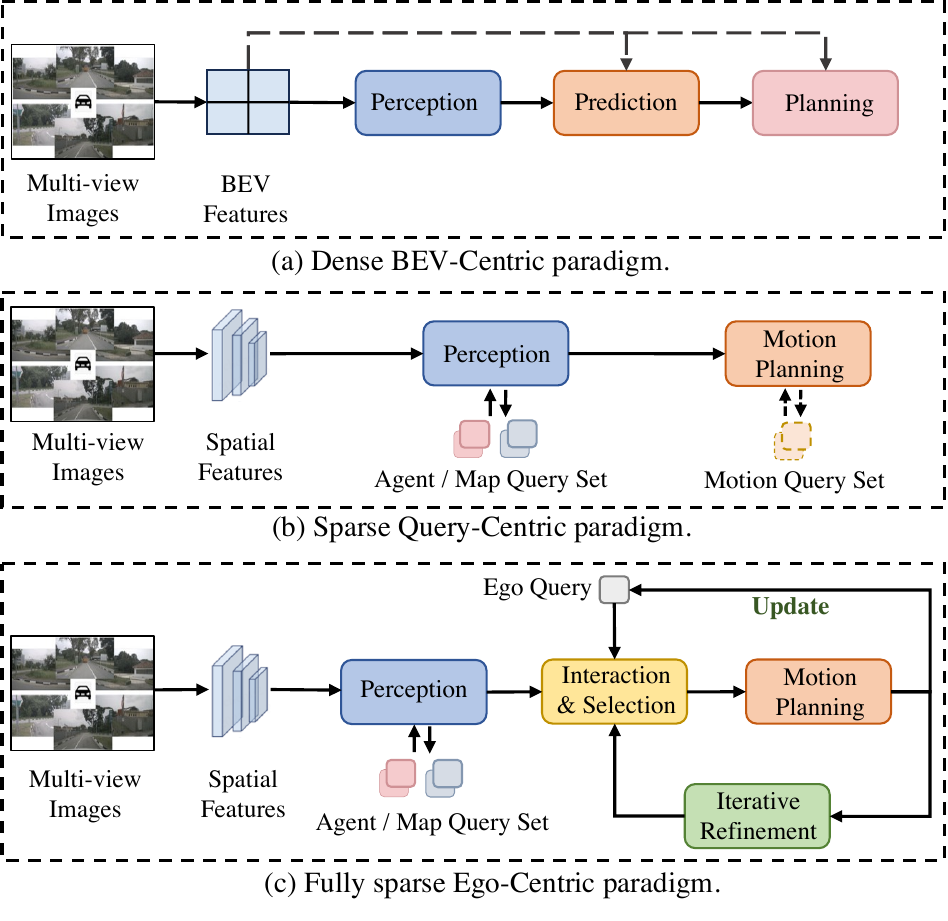}
\vspace{-0.2cm} 
\caption{Comparison of E2E-AD paradigms. (a) The dense \textbf{BEV-Centric} paradigm. (b) The sparse \textbf{Query-Centric} paradigm. (c) The proposed fully sparse \textbf{Ego-Centric} paradigm. 
}
\label{overview}
\vspace{-0.2cm} 
\end{figure}

Autonomous driving has experienced notable progress in recent years. Traditional driving systems are commonly decoupled into several standalone tasks, e.g. perception, prediction and planning. However, heavily relying on hand-crafted post-processing, the well-established modular systems suffer from information loss and error accumulation across sequential modules. Recently, end-to-end paradigm integrates all tasks into a unified model for planning-oriented optimization, showcasing great potential in pushing the limit of autonomous driving performance.

Literally, existing end-to-end models~\cite{hu2023planning,ye2023fusionad,jiang2023vad,sun2024sparsedrive, yang2023llm4drive, yang2025drivemoe, yang2025raw2drive} designed for reliable trajectory planning can be classified into two mainstreams as summarized in Fig.~\ref{overview}(a) and (b). The dense BEV-Centric paradigm~\cite{hu2023planning, ye2023fusionad} performs perception, prediction and planning consecutively upon the shared BEV (Bird's Eye View) features, which are computationally expensive leading to inferior efficiency. The sparse Query-Centric paradigm~\cite{sun2024sparsedrive} utilizes sparse representation to achieve scene understanding and joint motion planning, thus improving the overall efficiency. However, object-intensive motion prediction inevitably causes computational redundancy and violates the driving habits of human drivers, who usually only concentrate on the Closest In-Path Vehicle / Stationary (CIPV / CIPS) which are more likely to affect the driving intention and trajectory planning of ego-vehicle. Meanwhile, excessive interaction with irrelevant agents will be conversely adverse to the ego-planning. Therefore, the planning performance remains unsatisfactory in both planning safety, comfort and personification. 


To this end, we propose EgoFSD, an Ego-Centric fully sparse paradigm as shown in Fig.~\ref{overview}(c). Specifically, EgoFSD consists of sparse perception, hierarchical interaction and iterative motion planner. In the sparse perception module, multi-scale image features extracted from visual encoder are adopted for object detection and online mapping simultaneously in a sparse manner. Then the hierarchical interaction performs ego-centric and object-centric dual interaction to select the CIPV / CIPS with the help of an additional geometric prior. Thus the interactive queries can be selected gradually from coarse to fine. As for the motion planner, the mutual information between sparse interactive queries and ego-query is considered for motion prediction in a joint decoder, which is neglected in previous methods~\cite{hu2023planning,jiang2023vad} but is essential especially in the scenarios like intersections. To ensure the planning rationality and selection accuracy of interactive queries, the iterative planning optimization is further applied to the multi-modal proposal ego-trajectories, through continually updating the reference line and ego-query. Moreover, both position-level motion diffusion and trajectory-level planning denoising are introduced for stable training and fast convergence. It can not only model the uncertain positions of interactive agents for motion prediction, but also enhance the quality of trajectory refinement with arbitrary offsets. With above elaborate designs, EgoFSD exhibits the great potential of fully sparse paradigm for end-to-end autonomous driving, which significantly reduces the average L2 error by \textbf{59\%} and collision rate by \textbf{92\%} than UniAD~\cite{hu2023planning} respectively. Notably, our EgoFSD-B achieves $\textbf{6.9}\times$ faster running efficiency as well. In sum, the main contributions of our work are as follows:

\begin{itemize}[noitemsep,topsep=0pt,leftmargin=*,itemsep=2pt]
    \item We propose an \textbf{Ego-Centric Fully Sparse} paradigm for end-to-end self-\textbf{D}riving, named \textbf{EgoFSD}, without dense representation learning and redundant environmental modeling, which is proven to be effective for ego-planning.

    \item We introduce a geometric prior through intention-guided attention, where the \textbf{Closest In-Path Vehicle / Stationary (CIPV / CIPS)} are gradually picked out through ego-centric cross attention and selection. Besides, both \textbf{position-level diffusion} of interactive agents and \textbf{trajectory-level denoising} of ego-vehicle are adopted for uncertainty motion modeling.
  
    \item Extensive experiments are conducted on nuScenes~\cite{caesar2020nuscenes} and Bench2Drive~\cite{jia2024bench2drive} for planning evaluation, which demonstrate the superiority and prominent efficiency of our EgoFSD, revealing the great potential of the proposed ego-centric fully sparse paradigm.
\end{itemize}

\section{Related Work}
\label{sec:formatting}

\begin{figure*}
\centering
\includegraphics[width=18cm]{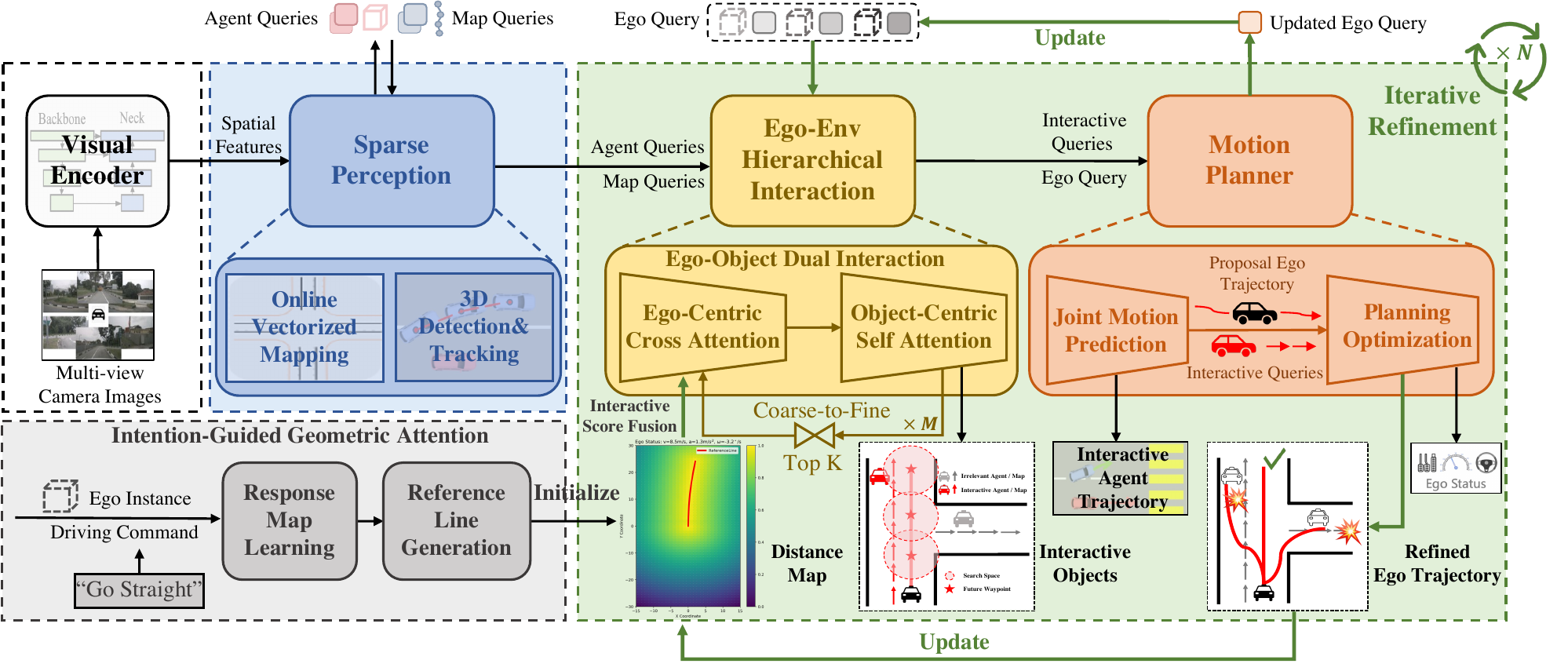}
\vspace{-0.6cm}
\caption{Overview of our proposed framework. EgoFSD first extracts multi-scale image features from multi-view images using an off-the-shelf visual encoder, then perceives both dynamic and static elements in a sparse manner. The Ego-Env hierarchical interaction module is presented to select the interactive queries from coarse to fine using three different driving commands of ego queries, which are leveraged for joint motion planner through iterative refinement. An additional geometric prior is introduced for high-quality query ranking through intention-guided attention. Besides, both position-level agent diffusion and trajectory-level ego-vehicle denoising are conducted for uncertainty modeling of the end-to-end driving system.}
\label{framework}
\vspace{-0.3cm}
\end{figure*}

\noindent
\textbf{Object Detection.} Recent years witness remarkable progress achieved in multi-view 3D detection, which mainly build elaborate designs upon the dense BEV (Bird's Eye View) or sparse query features. To generate BEV features, LSS~\cite{philion2020lift} lifts 2D image features to 3D space using depth estimation results, which are then splatted into BEV plane. Follow-up works apply such operation to perform view transform for 3D detection task~\cite{huang2021bevdet,li2023bevdepth,han2024exploring,jin2025unimamba,jin2025geoformer}. Differently, some works~\cite{li2022bevformer,yang2023bevformer,huang2023fast} project a series of predefined BEV queries in 3D space to the image domain for feature sampling. As for the sparse fashion, current methods~\cite{wang2022detr3d,liu2022petr,liu2023petrv2,liu2023sparsebev,lin2023sparse4d,su2024robosense,su2025freqpde} adopt a set of sparse queries to integrate spatial-temporal aggregations from multi-view image feature sequence for iterative anchor refinement.


\noindent
\textbf{Online Mapping.} Maps could provide important static scenario information to ensure driving safety. Current works~\cite{li2022hdmapnet,liu2023vectormapnet,liao2022maptr,yuan2024streammapnet} manage to construct online maps with on-board sensors, instead of using HD-Map which is labor intensive and expensive. HDMapNet~\cite{li2022hdmapnet} achieves this aim through BEV semantic segmentation and heuristic post-processing to generate map instances. VectorMapNet~\cite{liu2023vectormapnet} introduces a two-stage auto-regressive transformer to refine map elements consecutively. MapTR~\cite{liao2022maptr} regards map elements as a set of points with equivalent permutations, while StreamMapNet~\cite{yuan2024streammapnet} adopts a temporal fusion strategy to enhance the performance. However, all of them rely on dense BEV features for online map construction, which is computationally intensive and thus inefficient.  

\noindent
\textbf{Motion Prediction.} Predicting agent future trajectories is essential for the autonomous vehicle to understand motion intention of surrounding agents. FaF~\cite{luo2018fast} predicts both current and future boxes from images using a single deep network. IntentNet~\cite{casas2018intentnet} attempts to reason high-level behavior and long-term trajectories simultaneously. PnPNet~\cite{liang2020pnpnet} aggregate trajectory-level features for motion prediction through an online tracking module. ViP3D~\cite{gu2023vip3d} takes images and HD-Map as input, and adopts agent queries to conduct tracking and prediction. PIP~\cite{jiang2022perceive} further proposes to replace HD-Map with local vectorized map.

\noindent
\textbf{End-to-End Planning.} End-to-end planning paradigm either unites modules of perception and prediction~\cite{jiang2023vad,zhang2024graphad,ye2023fusionad, niu2023lightzero, pu2024unizero}, or adopts a direct optimization on planning without intermediate tasks~\cite{codevilla2018end,codevilla2019exploring, li2023normalization}, which lack interpretability and are hard to optimize. Recently, UniAD~\cite{hu2023planning} presents a planning-oriented model which integrates various tasks in the dense BEV-Centric paradigm, achieving convincing performance. VAD~\cite{jiang2023vad} learns vectorized scene representations and improves planning safety with explicit constraints. GraphAD~\cite{zhang2024graphad} constructs the interaction scene graph to model both dynamic and static relations. SparseDrive~\cite{sun2024sparsedrive} introduces the symmetric sparse perception for parallel motion planner, which consumes more computation cost due to the repeated query projection and deformable feature aggregation, without a shared 3D feature~\cite{hu2023planning,jiang2023vad}. Besides, using straightforward designs and exhaustive modeling without ego-centric interaction, will inevitably lead to unsatisfactory planning performance and inferior efficiency.

\section{Our Approach}

\subsection{Overview}
The overall framework of proposed EgoFSD is illustrated in Fig.~\ref{framework}, which deals with the end-to-end planning task in an ego-centric fully sparse paradigm. Specifically, EgoFSD mainly consists of four parts: visual encoder, sparse perception, hierarchical interaction and iterative motion planner. First, the visual encoder extracts multi-scale spatial features from given multi-view images. Then the sparse perception takes the encoded features as input to perform detection, tracking and online mapping simultaneously. In the hierarchical interaction module, the ego query equipped with a geometric prior is introduced to pick out the interactive queries through ego-centric cross attention and hierarchical selection. In the iterative motion planner, both interactive agents and ego-vehicle are considered for joint motion prediction, then the predicted multi-modal ego-trajectories are further optimized iteratively. Meanwhile, both position-level diffusion of interactive agents and trajectory-level denoising of ego-vehicle are conducted for uncertainty modeling of motion and planning tasks respectively.

\vspace{-0.1cm}
\subsection{Problem Formulation} 
Given multi-view camera image sequence can be denoted as $\mathbf{S}=\{I_{t} \in \mathbb{R}^{N\times 3\times H \times W}\}_{t=T-k}^{T}$, where $N$ is the number of camera views and $k$ indicates the temporal length till current timestep $T$ respectively. Annotation of input $\mathbf{S}$ for end-to-end planning is composed by a set of future waypoints of the ego-vehicle $\mathbf{\psi} = \{\phi=(x_t, y_t)\}_{t=1}^{T_{p}}$, where $T_{p}$ = 3$s$ is the planning time horizon, and ($x_t$, $y_t$) is the BEV location transformed to the ego-vehicle coordinate system at current timestep $T$. Meanwhile, driving command as well as ego-status is also provided. Annotation set $\mathbf{\psi}$ is used during training. During prediction, the planned trajectory of ego-vehicle should fit the annotation $\mathbf{\psi}$ with minimum L2 errors and collision rate with surrounding agents.



\vspace{-0.1cm}
\subsection{Sparse Perception}
After extracting the multi-view visual features $\mathbf{F}$ from sensor images using the visual encoder~\cite{he2016deep}, sparse query-based perception method proposed in~\cite{lin2023sparse4d,liu2022petr,wang2023exploring} is extended to perform unified detection and online mapping in parallel with the symmetric architecture as adopted in~\cite{sun2024sparsedrive}.

\noindent
\textbf{Detection.} Following~\cite{lin2023sparse4d,liu2023sparsebev}, surrounding agents can be represented by a group of instance features $\mathbf{F}_{a} \in \mathbb{R}^{N_{a}\times C}$ and anchor boxes $\mathbf{B}_{a} \in \mathbb{R}^{N_{a}\times11}$ respectively. And each anchor box $b_{a}$ can be denoted:

{\fontsize{8.5}{1}\selectfont
\begin{equation}
    b_{a} = \{x,y,z,ln(w),ln(h),ln (l),sin(\theta),con(\theta),v_{x},v_{y},v_{z}\},
\end{equation}}which contains location, dimension, yaw angle as well as velocity respectively. Taking $\mathbf{F}_{a}$, $\mathbf{B}_{a}$ and the visual features $\mathbf{F}$ as input, $N_{dec}$ decoders are adopted to consecutively refine the anchor boxes and update the instance features through iterative decoding. The updated instance features are adopted to predict the classification scores and box offsets respectively. Refer to~\cite{wang2023exploring}, hybrid attention between current propagated queries and history memory queue is conducted for temporal modeling. Besides, temporal instance denoising is introduced to improve model stability. 





\noindent
\textbf{Online Mapping.} Similarly, we adopt an additional detection branch of same structure for online mapping. Differently, the geometric anchor of each static map element is denoted as $N_{p}$ points. Therefore, surrounding maps can be represented by a group of map instance features $\mathbf{F}_{m} \in \mathbb{R}^{N_{m}\times C}$ and anchor polylines $\mathbf{B}_{m} \in \mathbb{R}^{N_{m}\times N_{p}\times 2}$.

\subsection{Ego-Env Hierarchical Interaction}
We continue to perform hierarchical interaction between the ego-vehicle and surrounding objects. As shown in Fig.~\ref{framework}, the hierarchical interaction module mainly consists of three parts: \textit{Ego-Object Dual Interaction}, \textit{Intention-guided Geometric Attention} and \textit{Coarse-to-Fine Selection}.

\begin{figure}[tb!]  
\centering
\includegraphics[width=7.8cm]{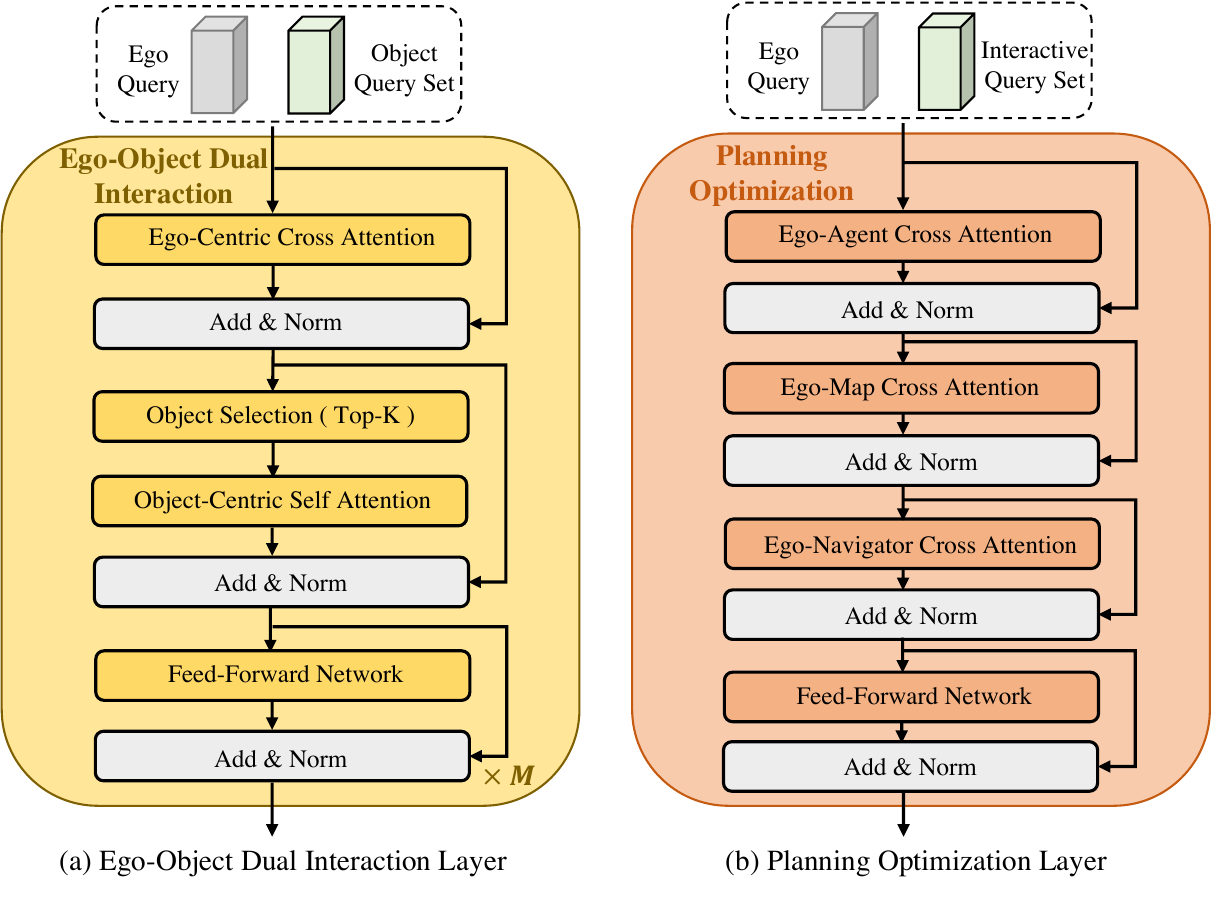}
\caption{Illustration of the \textbf{Dual Interaction} layer in the hierarchical interaction and \textbf{Planning Optimization} layer in the motion planner.}
\label{fig:interaction_planner}
\vspace{-5pt}
\end{figure}

\noindent
\textbf{Ego-Object Dual Interaction.} As shown in Fig.~\ref{fig:interaction_planner}, a learnable embedding $\mathbf{F}_{e} \in \mathbb{R}^{1 \times C}$ is randomly initialized to serve as ego query, along with an ego anchor box $\mathbf{B}_{e} \in \mathbb{R}^{1\times11}$ together to represent the ego-vehicle. Both ego-centric cross attention with surrounding objects $\mathbf{F}_{o} \in \mathbb{R}^{N_{o} \times C}$ ($N_{o} = N_{a} + N_{m})$ and object-centric self attention are conducted consecutively to capture the mutual information comprehensively. During the attention calculation process, we combine positional embedding and query feature in a concatenated manner instead of an additive approach, which can effectively retain both semantic and geometric clues for interaction modeling.

\noindent
\textbf{Intention-Guided Geometric Attention.} To enhance the accuracy and explainability of query ranking to facilitate selection, we introduce an ego-centric geometric prior additionally. As shown in Fig.~\ref{framework}, the intention-guided attention module is adopted to assess the importance of surrounding agent and map queries, which mainly consists of three steps: \textbf{response map learning}, \textbf{reference line generation} and \textbf{interactive score fusion}. 

Specifically, we use four MLPs to encode the ego-intention respectively, including velocity, acceleration, angular velocity and driving command. Then we concatenate these embeddings to obtain ego-intention features $\mathbf{I}_{e}\in \mathbb{R}^{1 \times C}$, which are further concatenated with the position embeddings $ \mathbf{F}_{p}\in \mathbb{R}^{H\times W\times C}$ of a group of pre-defined locations $\mathbf{P}\in \mathbb{R}^{H\times W\times2}$ to cover densely distributed grid cells in the BEV plane. Finally, the concatenated geometric features are fed to a single SE~\cite{hu2018squeeze} block to learn response map $\mathbf{M}_{r}\in \mathbb{R}^{H\times W\times 1}$, which is supervised by the normalized minimum distance from $\mathbf{P}$ to interpolated future trajectory $\psi'$ with $T_{e}'$ waypoints. Thus, for each grid $i$, the response value $M_{r}^{i}$ can be formulated as:

{\fontsize{7.3}{1}\selectfont
\begin{equation}
M_{r}^{i}=1- \min(\{||\mathbf{P}_{i} - \mathbf{\psi}_{j}'||_{2}\}_{j=1}^{T_{e}'}) / \max(\{\min(||\mathbf{P}_{i}-\mathbf{\psi}'||_{2})\}_{i=1}^{H\times W},
\end{equation}}where $T_{e}'$ is set to 30 for trajectory interpolation empirically. \textit{The motivation is that the Closest In-Path Vehicle / Stationary are prone to affect the ego-intention, and vice versa}.

With the predicted $\mathbf{M}_{r}$, we first generate the reference line through row-wise thresholding, which are further used to generate the normalized distance map $\mathbf{M}_{d}$ (See Fig.~\ref{framework}). Then we can obtain the geometric score $\mathbf{S}_{geo}$ for each object query by referring to the $\mathbf{M}_{d}$.
\textbf{The reason why we don't get the geometric score from $\mathbf{M}_{r}$ directly is that the imbalanced distribution of ego-intention and future waypoints~\cite{li2024ego} may lead to the inferior quality of $\mathbf{M}_{r}$.}

\begin{figure}[tb!]  
\includegraphics[width=8.5cm]{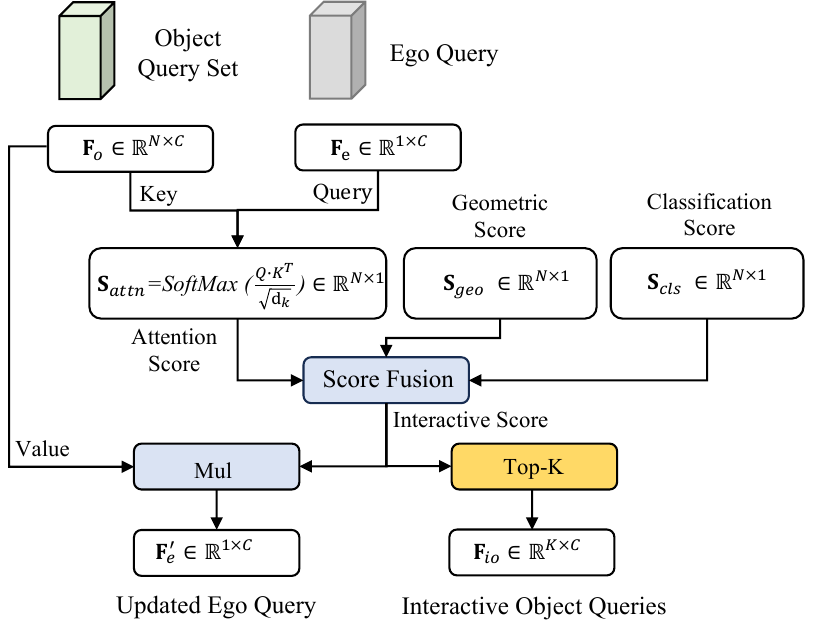}
\caption{Details of the \textbf{Interactive Score Fusion} process used for the geometric attended query selection.}
\label{fig:cross_attn}
\vspace{-5pt}
\end{figure}

Finally, as shown in Fig.~\ref{fig:cross_attn}, we perform interactive score fusion through multiplying the attention, geometric and classification scores during the ego-centric cross attention: 
\begin{equation}
    \mathbf{S}_{inter} = \underbrace{Softmax(\mathbf{F}_{e} \cdot \mathbf{F}_{o}^{T}/\sqrt{d_{k}})}_{\mathbf{S}_{attn}\in \mathbb{R}^{N\times 1}} \odot \mathbf{S}_{geo} \odot \mathbf{S}_{cls},
\label{eq1}
\end{equation}
where the distance-prior is weighted with the attention score $\mathbf{S}_{attn}$ for both interaction and selection. $\cdot$ is inner product, $\odot$ is Hadamard product, and $d_{k}$ is the channel dimension.


\noindent
\textbf{Coarse-to-Fine Selection.} To capture the interaction information from coarse to fine, we stack $M$ dual-interaction layers in a cascaded manner, where a top-$K$ operation is appended between each two consecutive layers, thus the interactive objects can be gradually selected for latter prediction and planning usages. We claim that only a few interactive objects need to be considered for motion prediction, which are enough yet efficient for ego-centric path planning, instead of all detected agents existing in the driving scene. $M$ is set to 3 and $K = \{10\%, 5\%, 2\%\}$ empirically.

\subsection{Iterative Motion Planner}
As shown in Fig.~\ref{framework}, the iterative motion planner is designed to conduct motion prediction for both interactive agents and ego-vehicle, and then optimize the ego-trajectory with both safety and kinematic constrains iteratively.

\noindent
\textbf{Joint Motion Prediction.} With regard to the trajectory prediction, both surrounding agents and ego-vehicle are adopted for motion modeling in a joint decoder, unlike previous works~\cite{hu2023planning,jiang2023vad,ye2023fusionad} which neglect the crucial interactions between near agents and ego-vehicle when making motion predictions, especially in the common scenarios like intersections. Another difference is that only the interactive objects $\mathbf{F}_{io}$ (CIPV) sparsely selected in the former module are considered, instead of all detected agents in the driving scene which maybe irrelevant to the ego-vehicle planning. As for the joint motion decoder, we prepare three copies of ego query $\mathbf{F}_{e}^{'}$ to indicate different driving intentions (\textit{i.e.,} turn left, turn right and keep forward), which are combined with $\mathbf{F}_{io}$ to conduct agent-level self-attention and agent-map cross attention respectively. And then we concatenate these enhanced features to predict multi-modal trajectories $\mathbf{\tau}_{a}\in \mathbb{R}^{N_{a}\times K_{a}\times T_{a}\times 2}$, $\mathbf{\tau}_{e}\in \mathbb{R}^{N_{e}\times K_{e}\times T_{e}\times 2}$ and classification scores $\mathbf{S}_{a}\in \mathbb{R}^{N_{a}\times K_{a}}$, $\mathbf{S}_{e}\in \mathbb{R}^{N_{e}\times K_{e}}$ for both agents and ego-vehicle, where $N_{e}=3$ is the number of driving command for planning, $K_{a}=K_{e}=6$ are the mode number, $T_{a}=T_{e}=6$ are the future timestamps.

\noindent
\textbf{Planning Optimization.} With the predicted multi-intention and multi-modal trajectories of ego-vehicle, we can select the most probable proposal trajectory with the input driving command and classification score $\mathbf{S}_{e}$. As shown in Fig.~\ref{fig:interaction_planner}(b), ego-agent, ego-map and ego-navigator cross attentions are conducted consecutively for planning optimization. And the offsets for each future waypoint are predicted upon the proposal trajectory respectively with several planning constraints proposed in \cite{jiang2023vad} to ensure safety.

\noindent
\textbf{Iterative Refinement.} To ensure the interaction quality and selection accuracy of interactive queries, an additional iterative refinement strategy is proposed to continuously update the reference line and distance map $\mathbf{M}_{d}$ with refined ego trajectory as illustrated in Fig.~\ref{framework}. 


\subsection{Uncertainty Denoising} 
Due to the planning-oriented modular design, output uncertainty from each individual module will be inevitably introduced and passed through to the downstream tasks, leading to inferior and fragile system. Under this circumstance, we propose a two-level uncertainty modeling strategy to further stabilize the whole framework. 

On one hand, \textbf{position-level diffusion process} is performed on Top-$K$ boxes of interactive agents $\mathbf{B}_{i} \in \mathbb{R}^{K\times 11}$ for additional trajectory prediction of noisy agents:
\begin{equation}
    \mathbf{B}_{n} = \mathbf{B}_{i} + \Delta \mathbf{B}_{pos} \in \mathbb{R}^{G\times K\times11},
\end{equation}
which are equipped with $G$ groups of random noises following uniform distributions. $\Delta B_{pos}$ locates within two different ranges of $\{-s, s\}$ and $\{-2s, -s\}\cup\{s, 2s\}$ to indicate positives and negatives respectively, where $s$ indicates the noise scale. This process aims to promote the stability of motion forecasting for interactive agents with uncertain detected positions, scales and velocities. 

On the other hand, \textbf{trajectory-level denoising process} is also introduced for robust offset prediction of proposal trajectory of ego-vehicle in the planning optimization stage. Different from the position diffusion of agent query on the purpose of detection and motion, we apply the random noise $\Delta \psi_{traj}$ to the ground-truth trajectory $\mathbf{\psi}_{gt}$ of ego-vehicle:
\begin{equation}
    \mathbf{\psi}_{n} = \mathbf{\psi}_{gt} + \Delta \mathbf{\psi}_{traj} \in \mathbb{R}^{G\times T_{e}\times2},
\end{equation}
where $\mathbf{\psi}_{n}$ indicate noisy trajectory proposals and the noise scale $s$ depends on the Final Displacement (FD) of $\mathbf{\psi}_{gt}$.



\begin{table*}[t]\footnotesize
\begin{center}
\caption{\textbf{Open-Loop Planning Evaluation Results on the nuScenes val Dataset}. $\ast$: LiDAR-based method. $\dag$: Reproduced with official checkpoint. $\ddagger$: Using evaluation protocol proposed in~\cite{li2024ego,zhai2023rethinking}.}
\label{table_open_loop}
\vspace{-0.2cm}
\setlength{\tabcolsep}{0.18cm}
\begin{tabular}{c|l|c|cccc|cccc|c|c}
\toprule[1pt]
\multirow{2}{*}{\textbf{Protocol}} & \multirow{2}{*}{\textbf{Method}} & \multirow{2}{*}{\textbf{Backbone}} & \multicolumn{4}{c|}{\textbf{L2 (\textit{m}) $\downarrow$}} & \multicolumn{4}{c|}{\textbf{Collision (\%)} $\downarrow$} & \multirow{2}{*}{\makecell{\textbf{Latency} \\ \textbf{(\textit{ms}) $\downarrow$}}} & \multirow{2}{*}{\textbf{FPS} $\uparrow$} \\ 

 & & & \multicolumn{1}{c}{1\textit{s}} & \multicolumn{1}{c}{2\textit{s}} & \multicolumn{1}{c}{3\textit{s}} & \cellcolor[HTML]{DADADA}Avg. & \multicolumn{1}{c}{1\textit{s}} & \multicolumn{1}{c}{2\textit{s}} & \multicolumn{1}{c}{3\textit{s}} & \cellcolor[HTML]{DADADA}Avg. &  &  \\ \midrule

\multirow{7}{*}{\makecell{\textbf{ST-P3} \\ \textbf{Metrics}}} & FF$^{\ast}$~\cite{hu2021safe} & - & \multicolumn{1}{c}{0.55} & \multicolumn{1}{c}{1.20} & \multicolumn{1}{c}{2.54} & \cellcolor[HTML]{DADADA}1.43 & \multicolumn{1}{c}{0.06} & \multicolumn{1}{c}{0.17} & \multicolumn{1}{c}{1.07} & \cellcolor[HTML]{DADADA}0.43 & - & - \\ 

& EO$^{\ast}$~\cite{khurana2022differentiable} & - & \multicolumn{1}{c}{0.67} & \multicolumn{1}{c}{1.36} & \multicolumn{1}{c}{2.78} & \cellcolor[HTML]{DADADA}1.60 & \multicolumn{1}{c}{0.04} & \multicolumn{1}{c}{0.09} & \multicolumn{1}{c}{0.88} & \cellcolor[HTML]{DADADA}0.33 & - & - \\ 

& ST-P3~\cite{hu2022st} & EfficientNet-b4 & \multicolumn{1}{c}{1.33} & \multicolumn{1}{c}{2.11} & \multicolumn{1}{c}{2.90} & \cellcolor[HTML]{DADADA}2.11 & \multicolumn{1}{c}{0.23} & \multicolumn{1}{c}{0.62} & \multicolumn{1}{c}{1.27} & \cellcolor[HTML]{DADADA}0.71 & 628.3 & 1.6 \\ 



 
& VAD-Base~\cite{jiang2023vad} & ResNet50 &  \multicolumn{1}{c}{0.41} & \multicolumn{1}{c}{0.70} & \multicolumn{1}{c}{1.05} & \cellcolor[HTML]{DADADA}0.72 & \multicolumn{1}{c}{0.07} & \multicolumn{1}{c}{0.17} & \multicolumn{1}{c}{0.41} & \cellcolor[HTML]{DADADA}0.22 & 224.3 & 4.5 \\ 

& SparseDrive-S~\cite{sun2024sparsedrive}$^{\dag}$ & ResNet50 &  \multicolumn{1}{c}{0.30} & \multicolumn{1}{c}{0.58} & \multicolumn{1}{c}{0.95} & \cellcolor[HTML]{DADADA}0.61 & \multicolumn{1}{c}{0.47} & \multicolumn{1}{c}{0.47} & \multicolumn{1}{c}{0.69} & \cellcolor[HTML]{DADADA}0.54 & 111.1 & 9.0\\ \cmidrule{2-13}
&  EgoFSD-S (BEV) & ResNet50 & \multicolumn{1}{c}{\textbf{0.16}} & \multicolumn{1}{c}{\textbf{0.33}}  & \multicolumn{1}{c}{\textbf{0.59}} & \cellcolor[HTML]{DADADA}\textbf{0.35} & \multicolumn{1}{c}{\textbf{0.00}} &  \multicolumn{1}{c}{\textbf{0.04}} & \multicolumn{1}{c}{\textbf{0.18}} & \cellcolor[HTML]{DADADA}\textbf{0.07} & 67.7 & 14.8 \\
 \midrule
 \multirow{7}{*}{\makecell{\textbf{SparseDrive} \\ \textbf{Metrics$^{\ddagger}$}}} 
 & UniAD~\cite{hu2023planning}$^{\dag}$ &  ResNet101-DCN & \multicolumn{1}{c}{0.45} & \multicolumn{1}{c}{0.70} & \multicolumn{1}{c}{1.04} & \cellcolor[HTML]{DADADA}0.73 & \multicolumn{1}{c}{0.62} & \multicolumn{1}{c}{0.58} & \multicolumn{1}{c}{0.63} & \cellcolor[HTML]{DADADA}0.61 & 555.6 & 1.8 \\ 

 & VAD-Base~\cite{jiang2023vad}$^{\dag}$ & ResNet50 &  \multicolumn{1}{c}{0.41} & \multicolumn{1}{c}{0.70} & \multicolumn{1}{c}{1.05} & \cellcolor[HTML]{DADADA}0.72 & \multicolumn{1}{c}{0.03} & \multicolumn{1}{c}{0.19} & \multicolumn{1}{c}{0.43} & \cellcolor[HTML]{DADADA}0.21 & 224.3 & 4.5 \\ 

 &  SparseDrive-S~\cite{sun2024sparsedrive} & ResNet50 &  \multicolumn{1}{c}{0.29} & \multicolumn{1}{c}{0.58} & \multicolumn{1}{c}{0.96} & \cellcolor[HTML]{DADADA}0.61 & \multicolumn{1}{c}{0.01} & \multicolumn{1}{c}{0.05} & \multicolumn{1}{c}{0.18} & \cellcolor[HTML]{DADADA}0.08 & 111.1 & 9.0 \\ 
 
 & SparseDrive-B~\cite{sun2024sparsedrive} & ResNet101 &  \multicolumn{1}{c}{0.29} & \multicolumn{1}{c}{0.55} & \multicolumn{1}{c}{0.91} & \cellcolor[HTML]{DADADA}0.58 & \multicolumn{1}{c}{0.01} & \multicolumn{1}{c}{\textbf{0.02}} & \multicolumn{1}{c}{\textbf{0.13}} & \cellcolor[HTML]{DADADA}\textbf{0.06} & 137.0 & 7.3 \\ \cmidrule{2-13}   

 & EgoFSD-S (BEV) & ResNet50 & \multicolumn{1}{c}{0.16} & \multicolumn{1}{c}{0.33}  & \multicolumn{1}{c}{0.59} & \cellcolor[HTML]{DADADA}0.35 & \multicolumn{1}{c}{0.03} &  \multicolumn{1}{c}{0.07} & \multicolumn{1}{c}{0.21} & \cellcolor[HTML]{DADADA}0.10 & 67.7 & 14.8 \\ 
 
 & EgoFSD-S (PV) & ResNet50 & \multicolumn{1}{c}{0.15} & \multicolumn{1}{c}{0.31}  & \multicolumn{1}{c}{0.56} & \cellcolor[HTML]{DADADA}0.33  & \multicolumn{1}{c}{\textbf{0.00}} &  \multicolumn{1}{c}{0.06} & \multicolumn{1}{c}{0.19} & \cellcolor[HTML]{DADADA}0.08 & \textbf{59.8} & \textbf{16.7} \\ 
 
 & EgoFSD-B (PV) & ResNet101 & \multicolumn{1}{c}{\textbf{0.12}} & \multicolumn{1}{c}{\textbf{0.28}}  & \multicolumn{1}{c}{\textbf{0.52}} & \cellcolor[HTML]{DADADA}\textbf{0.30}  & \multicolumn{1}{c}{\textbf{0.00}} &  \multicolumn{1}{c}{0.04} & \multicolumn{1}{c}{0.15} & \cellcolor[HTML]{DADADA}\textbf{0.06} & 80.5 & 12.4 \\ 
 \bottomrule[1pt]
\end{tabular}
\end{center}
\vspace{-0.7cm}
\end{table*}

\subsection{End-to-End Learning}
\noindent
\textbf{Multi-stage Training.} To facilitate the model convergence and training performance, we divide the training process into two stages. In stage-1, the sparse perception, hierarchical interaction and joint motion prediction tasks are trained from scratch to learn sparse scene representation, interaction and motion capability respectively. Note that no selection operation is adopted in stage-1, namely all detected agents are considered for motion forecasting to make full use of annotations. In stage-2, the geometric attention module and the iterative planning optimizer are added to train jointly for overall optimization with uncertainty modeling.

\noindent
\textbf{Loss Functions.} The overall optimization function mainly includes five tasks, where each task can be optimized with both classification and regression losses. The overall loss function for end-to-end training can be formulated as:
\begin{equation}
    \mathcal{L} = \mathcal{L}_{det} + \mathcal{L}_{map} + \mathcal{L}_{interact} + \sum_{i=1}^{N}(\mathcal{L}_{motion}^{i} + \mathcal{L}_{plan}^{i}),
\end{equation}
where $\mathcal{L}_{interact}$ is a combination of binary classification loss and $\mathcal{L}2$ regression loss to learn geometric score, where the positive (interactive) samples are denoted as grid cells with geometric score $\mathbf{S}_{geo} \geq 0.9$ (within 3$m$ for each future waypoint). An additional regression loss is included in $\mathcal{L}_{plan}$ for ego status prediction, instead of directly using it as input to the planner as~\cite{hu2023planning,ye2023fusionad,jiang2023vad}, which will lead to information leakage as proven in~\cite{li2024ego}. Meanwhile, vectorized planning constrains identified in~\cite{jiang2023vad} such as collision, overstepping and direction are also included in $\mathcal{L}_{plan}$ for regularization. Besides, the weight terms $\lambda_{task}$ of different losses are adjusted empirically to ensure the same magnitude. $N$ is the number of motion planning stages.

\section{Experiments}

\subsection{Datasets and Metrics}
Our experiments are first conducted on the challenging public nuScenes~\cite{caesar2020nuscenes} dataset, which contains 1000 driving scenes lasting 20 seconds respectively. Over 1.4M 3D bounding boxes of 23 categories are provided in total, which are annotated at 2Hz. Following the conventions~\cite{hu2023planning,jiang2023vad}, Collision Rate ($\%$) and L2 Displacement Error (DE) ($m$) are adopted to measure the open-loop planning performance. To study the effect of various perception encoders, we also evaluate the 3D object detection and online mapping results using mAP and NDS metrics respectively. Besides, Bench2Drive~\cite{jia2024bench2drive} provides a comprehensive benchmarking for evaluating multiple abilities of end-to-end AD systems in a closed-loop manner, which collects 1000 clips covering 44 interactive scenarios, 23 weathers and 12 towns in CARLA v2~\cite{dosovitskiy2017carla}. Following the official settings, we use 950 clips for training while leaving 50 clips for open-loop evaluation. As for the closed-loop evaluation, we run the trained model in CARLA with 220 test routes and calculate the closed-loop metrics such as Driving Score (DS), Success Rate (SR) and Efficiency, respectively.


\subsection{Implementation Details}

EgoFSD plans a 3 seconds future ego-trajectory with 2 seconds history information as input, which has two variants, namely EgoFSD-S and EgoFSD-B. As for EgoFSD-S, both Perspective-View version and BEV version are all implemented. ResNet50~\cite{he2016deep} is adopted as the default backbone network for visual encoding. The perception range is set to 60$m$$\times$30$m$ longitudinally and laterally. Input image size of EgoFSD-S is resized to  $256\times704$. For EgoFSD-S (BEV), $N_{dec}$ is 3, and the number of BEV query, map query, agent query are $100\times100$, $100\times20$ and 300, respectively. For EgoFSD-S (PV), $N_{a}$ is 900 and $N_{m}$ is 100 respectively. Each map element contains 20 map points. The feature dimension $C$ is 256. The noise scale $s$ is set to 2.0 and 0.2$\times$FD for motion and planning respectively. $G$ is set to 3. EgoFSD-B has deeper network (ResNet101, $N_{dec}$ = 6) and larger image resolution ($512\times1408)$. We use AdamW~\cite{loshchilov2017decoupled} optimizer and Cosine Annealing~\cite{loshchilov2016sgdr} scheduler to train EgoFSD with weight decay 0.01 and initial learning rate $2\times10$$^{-4}$. EgoFSD is trained for 60 epochs and 6 epochs on nuScenes and Bench2Drive respectively, running on 8 NVIDIA Tesla A800 GPUs with total batch size 32 empirically.



\begin{table}\footnotesize
\begin{center}
\caption{\textbf{Results on the Bench2Drive~\cite{jia2024bench2drive} Benchmark}. Both Open-Loop and Closed-Loop metrics are reported. Avg. L2 is averaged over future 2 seconds under 2Hz.} 
\label{table_closed_loop}
\vspace{-0.3cm}
\resizebox{1.0\columnwidth}{!}{
\begin{tabular}{l|c|ccc}
\toprule[1pt]
\multirow{2}{*}{\textbf{Method}} & \multicolumn{1}{c|}{\textbf{Open-loop Metric}} & \multicolumn{3}{c}{\textbf{Closed-loop Metrics}} \\ \cmidrule{2-5}
 &  \multicolumn{1}{c|}{Avg. L2 $\downarrow$ (\textit{m})} & \multicolumn{1}{c}{\cellcolor[HTML]{DADADA}Driving Score $\uparrow$}  & \multicolumn{1}{c}{Success Rate $\uparrow$ (\%)} & \multicolumn{1}{c}{Efficiency $\uparrow$}  \\ \midrule
 
 AD-MLP~\cite{zhai2023rethinking}  & \multicolumn{1}{c|}{3.64} & \multicolumn{1}{c}{\cellcolor[HTML]{DADADA}18.05} & \multicolumn{1}{c}{0.00} &  \multicolumn{1}{c}{48.45}  \\ 
 
 UniAD-Tiny~\cite{hu2023planning}  & \multicolumn{1}{c|}{0.80} & \multicolumn{1}{c}{\cellcolor[HTML]{DADADA}40.73} & \multicolumn{1}{c}{13.18} &  \multicolumn{1}{c}{123.92}  \\ 

 UniAD-Base~\cite{hu2023planning} & \multicolumn{1}{c|}{0.73} & \multicolumn{1}{c}{\cellcolor[HTML]{DADADA}45.81} & \multicolumn{1}{c}{16.36} &  \multicolumn{1}{c}{129.21}  \\ 
 
 VAD~\cite{jiang2023vad}  & \multicolumn{1}{c|}{0.91} & \multicolumn{1}{c}{\cellcolor[HTML]{DADADA}42.35} & \multicolumn{1}{c}{15.00} &  \multicolumn{1}{c}{157.94}  \\ 

 GenAD~\cite{zheng2024genad}   &  - & \multicolumn{1}{c}{\cellcolor[HTML]{DADADA}44.81} & 15.90 & - \\

 MomAD~\cite{song2025don}  & 0.82 & \multicolumn{1}{c}{\cellcolor[HTML]{DADADA}44.54} & 16.71 & 170.21 \\

 \midrule
 EgoFSD-S  & \multicolumn{1}{c|}{0.70} & \multicolumn{1}{c}{\cellcolor[HTML]{DADADA}52.02} & \multicolumn{1}{c}{21.00} &  \multicolumn{1}{c}{178.30} \\ 

 EgoFSD-B  & \multicolumn{1}{c|}{\textbf{0.66}} & \multicolumn{1}{c}{\textbf{\cellcolor[HTML]{DADADA}60.39}} & \multicolumn{1}{c}{\textbf{31.78}} & \multicolumn{1}{c}{\textbf{180.63}} \\

 \bottomrule[1pt]
\end{tabular}}
\end{center}
\vspace{-0.6cm}
\end{table}

\subsection{Main Results}
\noindent
\textbf{Open-Loop Planning Evaluation.} 
As shown in Tab.~\ref{table_open_loop}, EgoFSD demonstrates significant advantages in both performance and efficiency compared to previous works. On one hand, EgoFSD-S achieves the minimum L2 error despite its lightweight visual backbone and inferior BEV perception. Specifically, compared with BEVFormer-based end-to-end methods~\cite{hu2023planning,jiang2023vad}, EgoFSD-S (BEV) reduces the average L2 error by a great margin (0.38$m$ and 0.37$m$, separately), while significantly reducing the average collision rates by 84\% and 52\% respectively. Equipped with deeper visual backbone and advanced sparse detectors from Perspective View (PV), the average L2 error and collision rates can be further reduced to 0.30$m$ and to 0.06$\%$ respectively. On the other hand, benefiting from the ego-centric hierarchical interaction, \textit{only sparse interactive agents (2\%) are considered for motion planning}. Hence, EgoFSD-S can achieve great efficiency with 16.7 FPS, $9.3\times$ and $3.7\times$ faster than UniAD~\cite{hu2023planning} and VAD~\cite{jiang2023vad} respectively.

\noindent
\textbf{Closed-Loop Planning Evaluation.} We further validate the closed-loop performance in Bench2Drive~\cite{jia2024bench2drive}, which has been proposed recently for comprehensive benchmarking of end-to-end planning methods. As shown in Tab.~\ref{table_closed_loop}, AD-MLP~\cite{zhai2023rethinking} has a high L2 error and bad closed-loop planning performance using merely ego status as input, which is different from findings in nuScenes~\cite{caesar2020nuscenes}, demonstrating the behavior diversity in Bench2Drive. UniAD~\cite{hu2023planning} has a lower L2 error compared to VAD~\cite{jiang2023vad} but with worse closed-loop planning performance as discussed in~\cite{li2024ego}. Notably, EgoFSD-S achieves both the lowest L2 error 
and best closed-loop performance with great efficiency, showcasing the superiority and generalizability of our proposed method.


\subsection{Ablation Study}
\label{sec:ablation}

We conduct extensive experiments to study the effectiveness of our EgoFSD. We use EgoFSD-S as default for ablation. More experiments are provided in the Appendix.

\noindent
\textbf{Necessity of Geometric Prior.} We claim that the Closest In-Path Vehicle as well as Stationary (CIPV / CIPS) are more likely to interact with the ego-vehicle. To verify the necessity of such geometric prior, we conduct exhaustive ablations of the ego-centric query selector as show in Tab.~\ref{table_geometry}. Without ego-centric selection, fewer objects randomly selected can result in worse planning results. While using the ego-centric cross attention, only 2\% of surrounding queries are enough for achieving convincing planning performance. Besides, introducing the geometric prior through attention can dramatically reduce the collision rate by 42\% especially, thanks to the superior interactive planning. Meanwhile, when utilizing the Ground-Truth geometric score for upper-limit evaluation, we can obtain the extremely excellent planning performance (0.23$m$ average L2 error). \textbf{The proposed ego-centric query selector equipped with geometric attention is nontrivial for motion planner.}





\noindent
\textbf{Effect of designs in Hierarchical Interaction.} Tab.~\ref{table_interaction} shows the effectiveness of our elaborate designs in the hierarchical interaction module, which contains three main designs such as Dual Interaction (DI), Geometric Attention (GA) and Coarse-to-Fine Selection (CFS). DI models both ego-centric and object-centric interactions respectively, which ensures the query selection quality for interactive planning. GA facilitates the query selection process as discussed in Tab.~\ref{table_geometry}, which reduces the collision rate obviously owing to the specialized attention on CIPV / CIPS. And CFS contributes to the multi-granularity interaction modeling through hierarchical receptive fields from global to local. All of these three designs combined together can achieve overall convincing planning performance.

\begin{table}[tb!] \footnotesize
\begin{center}
\caption{Necessity of the ego-centric \textbf{Query Selector} and effect of the \textbf{Geometric Prior}.} 
\label{table_geometry}
\vspace{-0.3cm}
\resizebox{1.0\columnwidth}{!}{
\begin{tabular}{cc|cccc|cccc}
\toprule[1pt]
\multirow{2}{*}{\makecell{\textbf{Object} \\ \textbf{Selection}}} & \multirow{2}{*}{\makecell{\textbf{Geometric} \\  \textbf{Attention}}} & \multicolumn{4}{c|}{\textbf{Planning L2 (\textit{m}) $\downarrow$}} & \multicolumn{4}{c}{\textbf{Planning Coll. (\%)} $\downarrow$} \\ 
 & & \multicolumn{1}{c}{1\textit{s}} & \multicolumn{1}{c}{2\textit{s}} & \multicolumn{1}{c}{3\textit{s}} & \cellcolor[HTML]{DADADA}Avg. & \multicolumn{1}{c}{1\textit{s}} & \multicolumn{1}{c}{2\textit{s}} & \multicolumn{1}{c}{3\textit{s}} & \cellcolor[HTML]{DADADA}Avg. \\ \midrule
 
 100\% & \ding{55} & \multicolumn{1}{c}{0.27} & \multicolumn{1}{c}{0.47} & \multicolumn{1}{c}{0.74} & \cellcolor[HTML]{DADADA}0.49 & \multicolumn{1}{c}{0.10} & \multicolumn{1}{c}{0.21} & \multicolumn{1}{c}{0.37} & \cellcolor[HTML]{DADADA}0.22 \\ 
 
 Random (5\%) & \ding{55} & \multicolumn{1}{c}{0.28} & \multicolumn{1}{c}{0.49} & \multicolumn{1}{c}{0.79} & \cellcolor[HTML]{DADADA}0.52 & \multicolumn{1}{c}{0.08} & \multicolumn{1}{c}{0.17} & \multicolumn{1}{c}{0.38} & \cellcolor[HTML]{DADADA}0.21 \\ 

 Random (2\%) & \ding{55} & \multicolumn{1}{c}{0.33} & \multicolumn{1}{c}{0.57} & \multicolumn{1}{c}{0.87} & \cellcolor[HTML]{DADADA}0.59 & \multicolumn{1}{c}{0.18} & \multicolumn{1}{c}{0.30} & \multicolumn{1}{c}{0.51} & \cellcolor[HTML]{DADADA}0.33 \\

 0\% & \ding{55} & \multicolumn{1}{c}{2.25} & \multicolumn{1}{c}{3.75} & \multicolumn{1}{c}{5.26} & \cellcolor[HTML]{DADADA}3.75 & \multicolumn{1}{c}{2.82} & \multicolumn{1}{c}{5.42} & \multicolumn{1}{c}{6.39} & \cellcolor[HTML]{DADADA}4.88 \\ 



 Attn (5\%) & \ding{55} & \multicolumn{1}{c}{0.16} & \multicolumn{1}{c}{0.34} & \multicolumn{1}{c}{0.63} & \cellcolor[HTML]{DADADA}0.38 & \multicolumn{1}{c}{0.07} & \multicolumn{1}{c}{0.09} & \multicolumn{1}{c}{0.31} & \cellcolor[HTML]{DADADA}0.16 \\ 

 Attn (2\%) & \ding{55} & \multicolumn{1}{c}{0.16} & \multicolumn{1}{c}{0.34} & \multicolumn{1}{c}{0.61} & \cellcolor[HTML]{DADADA}\textbf{0.37} & \multicolumn{1}{c}{0.06} & \multicolumn{1}{c}{0.08} & \multicolumn{1}{c}{0.24} & \cellcolor[HTML]{DADADA}\textbf{0.12} \\ 
 
 \midrule
 Attn (2\%) & Random & \multicolumn{1}{c}{0.17} & \multicolumn{1}{c}{0.36} & \multicolumn{1}{c}{0.67} & \cellcolor[HTML]{DADADA}0.40 & \multicolumn{1}{c}{0.07} & \multicolumn{1}{c}{0.10} & \multicolumn{1}{c}{0.34} & \cellcolor[HTML]{DADADA}0.17 \\

 Attn (2\%) & GroundTruth & \multicolumn{1}{c}{0.14} & \multicolumn{1}{c}{0.23} & \multicolumn{1}{c}{0.33} & \cellcolor[HTML]{DADADA}\underline{0.23} & \multicolumn{1}{c}{0.07} & \multicolumn{1}{c}{0.08} & \multicolumn{1}{c}{0.10} & \cellcolor[HTML]{DADADA}\underline{0.07} \\

 Attn (2\%) & \checkmark & \multicolumn{1}{c}{0.16} & \multicolumn{1}{c}{0.33} & \multicolumn{1}{c}{0.59} & \cellcolor[HTML]{DADADA}\textbf{0.35} {\bf  \color{red}(-5\%)} & \multicolumn{1}{c}{0.00} & \multicolumn{1}{c}{0.04} & \multicolumn{1}{c}{0.18} & \cellcolor[HTML]{DADADA}\textbf{0.07} {\bf  \color{red}(-42\%)} \\
 \bottomrule[1pt]
\end{tabular}}
\end{center}
\vspace{-0.5cm}
\end{table}

\begin{table}[tb!]
\centering
\centering
\caption{Ablation for designs in the \textbf{Hierarchical Interaction}. ``DI": dual interaction; ``GA": geometric attention; ``CFS": coarse-to-fine selection.}
\label{table_interaction}
\vspace{-0.3cm}
\resizebox{1.0\columnwidth}{!}{
\begin{tabular}{ccc|cccc|cccc}
\toprule[1pt]
\multirow{2}{*}{\textbf{DI}} & \multirow{2}{*}{\textbf{GA}} & \multirow{2}{*}{\textbf{CFS}} & \multicolumn{4}{c|}{\textbf{Planning L2 (\textit{m}) $\downarrow$}} & \multicolumn{4}{c}{\textbf{Planning Coll. (\%)} $\downarrow$} \\ 
 & & & \multicolumn{1}{c}{1\textit{s}} & \multicolumn{1}{c}{2\textit{s}} & \multicolumn{1}{c}{3\textit{s}} & \cellcolor[HTML]{DADADA}Avg. & \multicolumn{1}{c}{1\textit{s}} & \multicolumn{1}{c}{2\textit{s}} & \multicolumn{1}{c}{3\textit{s}} & \cellcolor[HTML]{DADADA}Avg. \\ \midrule

 \ding{55}  & \checkmark & \checkmark & \multicolumn{1}{c}{0.16} & \multicolumn{1}{c}{0.33} & \multicolumn{1}{c}{0.59} & \cellcolor[HTML]{DADADA}0.36 & \multicolumn{1}{c}{0.01} & \multicolumn{1}{c}{0.08} & \multicolumn{1}{c}{0.23} & \cellcolor[HTML]{DADADA}0.11  \\ 
 
 \checkmark  & \ding{55} & \checkmark & \multicolumn{1}{c}{0.16} & \multicolumn{1}{c}{0.34} & \multicolumn{1}{c}{0.61} & \cellcolor[HTML]{DADADA}0.37 & \multicolumn{1}{c}{0.06} & \multicolumn{1}{c}{0.08} & \multicolumn{1}{c}{0.24} & \cellcolor[HTML]{DADADA}0.12 \\
 
 \checkmark  & \checkmark & \ding{55} & \multicolumn{1}{c}{0.19} & \multicolumn{1}{c}{0.37} & \multicolumn{1}{c}{0.64} & \cellcolor[HTML]{DADADA}0.40 & \multicolumn{1}{c}{0.09} & \multicolumn{1}{c}{0.12} & \multicolumn{1}{c}{0.23} & \cellcolor[HTML]{DADADA}0.14 \\ 
  
 \checkmark & \checkmark & \checkmark & \multicolumn{1}{c}{0.16} & \multicolumn{1}{c}{0.33} & \multicolumn{1}{c}{0.59} & \cellcolor[HTML]{DADADA}\textbf{0.35} & \multicolumn{1}{c}{0.00} & \multicolumn{1}{c}{0.04} & \multicolumn{1}{c}{0.18} & \cellcolor[HTML]{DADADA}\textbf{0.07} \\ 
 \bottomrule[1pt]
\end{tabular}}
\vspace{-0.1cm}
\end{table}

 \begin{table}[tb!]
\centering
\caption{Ablation for designs in the \textbf{Motion Planner}. ``JMP" means joint motion prediction; ``PO" means planning optimization; ``IR" means iterative refinement. ``UD" means uncertainty denoising.}
\label{table_motion}
\vspace{-0.3cm}
\resizebox{1.0\columnwidth}{!}{
\begin{tabular}{c|cccc|cccc|cccc}
\toprule[1pt]
\multirow{2}{*}{\textbf{ID}} & \multirow{2}{*}{\textbf{JMP}} & \multirow{2}{*}{\textbf{PO}} & \multirow{2}{*}{\textbf{IR}} & \multirow{2}{*}{\textbf{UD}} & \multicolumn{4}{c|}{\textbf{Planning L2 (\textit{m}) $\downarrow$}} & \multicolumn{4}{c}{\textbf{Planning Coll. (\%)} $\downarrow$} \\ 
 & & & & & \multicolumn{1}{c}{1\textit{s}} & \multicolumn{1}{c}{2\textit{s}} & \multicolumn{1}{c}{3\textit{s}} & \cellcolor[HTML]{DADADA}Avg. & \multicolumn{1}{c}{1\textit{s}} & \multicolumn{1}{c}{2\textit{s}} & \multicolumn{1}{c}{3\textit{s}} & \cellcolor[HTML]{DADADA}Avg. \\ \midrule
 
 1 & \checkmark & \ding{55} & \ding{55} & \checkmark & \multicolumn{1}{c}{0.23} & \multicolumn{1}{c}{0.48} & \multicolumn{1}{c}{0.83} & \cellcolor[HTML]{DADADA}0.51 & \multicolumn{1}{c}{0.08} & \multicolumn{1}{c}{0.13} & \multicolumn{1}{c}{0.35} & \cellcolor[HTML]{DADADA}0.18 \\ 
 
 2  & \checkmark & \checkmark & \ding{55} & \checkmark & \multicolumn{1}{c}{0.16} & \multicolumn{1}{c}{0.33} & \multicolumn{1}{c}{0.61} & \cellcolor[HTML]{DADADA}0.37 & \multicolumn{1}{c}{0.01} & \multicolumn{1}{c}{0.08} & \multicolumn{1}{c}{0.23} & \cellcolor[HTML]{DADADA}0.11 \\ 

 3  & \checkmark & \checkmark & \checkmark & \ding{55} & \multicolumn{1}{c}{0.16} & \multicolumn{1}{c}{0.34} & \multicolumn{1}{c}{0.64} & \cellcolor[HTML]{DADADA}0.38 & \multicolumn{1}{c}{0.07} & \multicolumn{1}{c}{0.07} & \multicolumn{1}{c}{0.17} & \cellcolor[HTML]{DADADA}0.10 \\ 

 4 & \checkmark & \checkmark & \checkmark & \checkmark & \multicolumn{1}{c}{0.16} & \multicolumn{1}{c}{0.33} & \multicolumn{1}{c}{0.59} & \cellcolor[HTML]{DADADA}\textbf{0.35} & \multicolumn{1}{c}{0.00} & \multicolumn{1}{c}{0.04} & \multicolumn{1}{c}{0.18} & \cellcolor[HTML]{DADADA}\textbf{0.07} \\ 
 \bottomrule[1pt]
\end{tabular}}
\vspace{-0.2cm}
\end{table}

\noindent
\textbf{Effect of designs in Motion Planner.} As for motion planner in EgoFSD, Joint Motion Prediction (JMP), Planning Optimization (PO) as well as Iterative Refinement (IR) makes up the planning pipeline of ego-vehicle. Besides, Uncertain Denoising (UD) contributes to the system stability and training convergence. Tab.~\ref{table_motion} explores the effect of each design exhaustively. ID-1 indicates evaluating the proposal trajectory of ego-vehicle predicted together with interactive agents, which achieves competitive L2 error but is easier to collide with surrounding agents. ID-2 improves the collision rate greatly by 38.9\% with the help of PO and planning constraints~\cite{jiang2023vad} during training phase. ID-4 emphasizes the importance of IR in improving the quality of ego-planning trajectory (average 5.4\% L2 error and 36.3\% collision rate reduction respectively). ID-3 reflects the benefit of UD used for end-to-end training compared to ID-4.

\noindent
\textbf{Effect of Iterative Refinement stages.} We continue to study the number of refinement stages in Tab.~\ref{tab_refinement}. We can observe that our EgoFSD can obtain superior planning performance with one additional refinement stage (36.3\% collision rate reduction), which becomes saturated when introducing more stages. Hence, two-stage interacted motion planner is enough for achieving convincing results.

\noindent
\textbf{Effect of Uncertainty Denoising.} We also validate the effectiveness of uncertainty denoising strategy including position-level motion diffusion and trajectory-level planning denoising. As shown in Tab.~\ref{tab_uncertainty}, motion diffusion can improve the prediction stability with uncertain agent positions, while the planning denoising can also strengthen the trajectory regression precision of ego-vehicle.

\begin{table}[tb!]\footnotesize
\centering
\caption{Ablation for \textbf{Iterative Refinement} stages.} \label{tab_refinement}
\vspace{-0.3cm}
\resizebox{1.0\columnwidth}{!}{
\begin{tabular}{c|cccc|cccc}
\toprule[1pt]
\multirow{2}{*}{\textbf{Number of Stages}}  & \multicolumn{4}{c|}{\textbf{Planning L2 (\textit{m}) $\downarrow$}} & \multicolumn{4}{c}{\textbf{Planning Coll. (\%)} $\downarrow$} \\ 
 & \multicolumn{1}{c}{1\textit{s}} & \multicolumn{1}{c}{2\textit{s}} & \multicolumn{1}{c}{3\textit{s}} & \cellcolor[HTML]{DADADA}Avg. & \multicolumn{1}{c}{1\textit{s}} & \multicolumn{1}{c}{2\textit{s}} & \multicolumn{1}{c}{3\textit{s}} & \cellcolor[HTML]{DADADA}Avg. \\ \midrule
  1 & \multicolumn{1}{c}{0.16} & \multicolumn{1}{c}{0.33} & \multicolumn{1}{c}{0.61} & \cellcolor[HTML]{DADADA}0.37 & \multicolumn{1}{c}{0.01} & \multicolumn{1}{c}{0.08} & \multicolumn{1}{c}{0.23} & \cellcolor[HTML]{DADADA}0.11 \\ 

  2 & \multicolumn{1}{c}{0.16} & \multicolumn{1}{c}{0.33} & \multicolumn{1}{c}{0.59} & \cellcolor[HTML]{DADADA}\textbf{0.35} & \multicolumn{1}{c}{0.00} & \multicolumn{1}{c}{0.04} & \multicolumn{1}{c}{0.18} & \cellcolor[HTML]{DADADA}\textbf{0.07} \\

  3 & \multicolumn{1}{c}{0.16} & \multicolumn{1}{c}{0.33} & \multicolumn{1}{c}{0.60} & \cellcolor[HTML]{DADADA}0.36 & \multicolumn{1}{c}{0.01} & \multicolumn{1}{c}{0.40} & \multicolumn{1}{c}{0.22} & \cellcolor[HTML]{DADADA}0.09 \\

  4 & \multicolumn{1}{c}{0.16} & \multicolumn{1}{c}{0.33} & \multicolumn{1}{c}{0.61} & \cellcolor[HTML]{DADADA}0.36 & \multicolumn{1}{c}{0.00} & \multicolumn{1}{c}{0.04} & \multicolumn{1}{c}{0.20} & \cellcolor[HTML]{DADADA}0.08 \\
 \bottomrule[1pt]
\end{tabular}}
\end{table}

\begin{table}[tb!]\footnotesize
\caption{Ablation for \textbf{Uncertainty Denoising} on both position and trajectory aspects.} \label{tab_uncertainty}
\vspace{-0.3cm}
\resizebox{1.0\columnwidth}{!}{
\begin{tabular}{cc|cccc|cccc}
\toprule[1pt]
\multirow{2}{*}{\makecell{\textbf{Position}\\ \textbf{Diffusion}}} & \multirow{2}{*}{\makecell{\textbf{Trajectory} \\ \textbf{Denoising}}}  & \multicolumn{4}{c|}{\textbf{Planning L2 (\textit{m}) $\downarrow$}} & \multicolumn{4}{c}{\textbf{Planning Coll. (\%)} $\downarrow$} \\ 
 & & \multicolumn{1}{c}{1\textit{s}} & \multicolumn{1}{c}{2\textit{s}} & \multicolumn{1}{c}{3\textit{s}} & \cellcolor[HTML]{DADADA}Avg. & \multicolumn{1}{c}{1\textit{s}} & \multicolumn{1}{c}{2\textit{s}} & \multicolumn{1}{c}{3\textit{s}} & \cellcolor[HTML]{DADADA}Avg. \\ \midrule
 

  \ding{55} & \ding{55} & \multicolumn{1}{c}{0.16} & \multicolumn{1}{c}{0.34} & \multicolumn{1}{c}{0.64} & \cellcolor[HTML]{DADADA}0.38 & \multicolumn{1}{c}{0.07} & \multicolumn{1}{c}{0.07} & \multicolumn{1}{c}{0.17} & \cellcolor[HTML]{DADADA}0.10 \\

  \checkmark & \ding{55} & \multicolumn{1}{c}{0.16} & \multicolumn{1}{c}{0.34} & \multicolumn{1}{c}{0.63} & \cellcolor[HTML]{DADADA}0.37 & \multicolumn{1}{c}{0.02} & \multicolumn{1}{c}{0.04} & \multicolumn{1}{c}{0.15} & \cellcolor[HTML]{DADADA}0.07 \\


  \checkmark & \checkmark & \multicolumn{1}{c}{0.16} & \multicolumn{1}{c}{0.33} & \multicolumn{1}{c}{0.59} & \cellcolor[HTML]{DADADA}\textbf{0.35} & \multicolumn{1}{c}{0.00} & \multicolumn{1}{c}{0.04} & \multicolumn{1}{c}{0.18} & \cellcolor[HTML]{DADADA}\textbf{0.07} \\
 \bottomrule[1pt]
\end{tabular}}
\vspace{-5pt}
\end{table}

\noindent
\textbf{Runtime of each module.} We evaluate the modular runtime of EgoFSD-S with Perspective-View perception paradigms as shown in Tab.~\ref{table_runtime}. Generally, visual backbone and perception decoder occupy the most of the runtime (57.7\%) for feature extraction and scene understanding. Hierarchical interaction also takes a significant part for interactive query selection. Thanks to the ego-centric interaction and selection module, the joint motion planner only consumes 7.5\textit{ms} to plan the future ego-trajectory. 
 

\subsection{Qualitative Results}

We visualize the motion trajectories of sparse interactive agents as well as planning results of EgoFSD as illustrated in Fig.~\ref{fig_visualization}. Both surrounding camera images and prediction results on BEV are provided accordingly. Besides, we also project the planning trajectories to the front-view camera image. Only the top-3 trajectories of selected agents interacting with ego-vehicle are visualized for better understanding of EgoFSD motivation. EgoFSD outputs planning results based on the vectorized representation in an end-to-end manner, not requiring any dense interaction and redundant motion modeling, let alone hand-crafted post-processing.

\begin{table}\footnotesize
\begin{center}
\caption{\textbf{Module Runtime Statistics}. The inference speed is measured for EgoFSD-S on NVIDIA GeForce RTX 3090 GPU as~\cite{jiang2023vad}.} 
\label{table_runtime}  
\vspace{-0.3cm}
\resizebox{1.0\linewidth}{!}{
\begin{tabular}{l|rrr}
\toprule[1pt]
\textbf{Module} &  \textbf{\#Param (M)} &  \textbf{Latency (\textit{ms})} & \textbf{Proportion (\%)} \\ \midrule
 Backbone & 26.7 & 4.7 & 7.9 \\ 
 Sparse Perception & 11.2 &  29.8 & 49.8 \\ 
 Hierarchical Interaction & 21.3 & 17.8 & 29.8 \\ 
 Joint Motion Prediction & 4.9 & 4.1 & 6.9\\ 
 \cellcolor[HTML]{DADADA}Planning Optimization & \cellcolor[HTML]{DADADA}1.8 & \cellcolor[HTML]{DADADA}3.4 & \cellcolor[HTML]{DADADA}5.6 \\ 
 \midrule
 Total & 65.9 & 59.8 & 100.0 \\ 
 \bottomrule[1pt]
\end{tabular}}
\end{center}
\vspace{-0.5cm}
\end{table}

\begin{figure}
	\centering
	\includegraphics[width=0.95\linewidth]{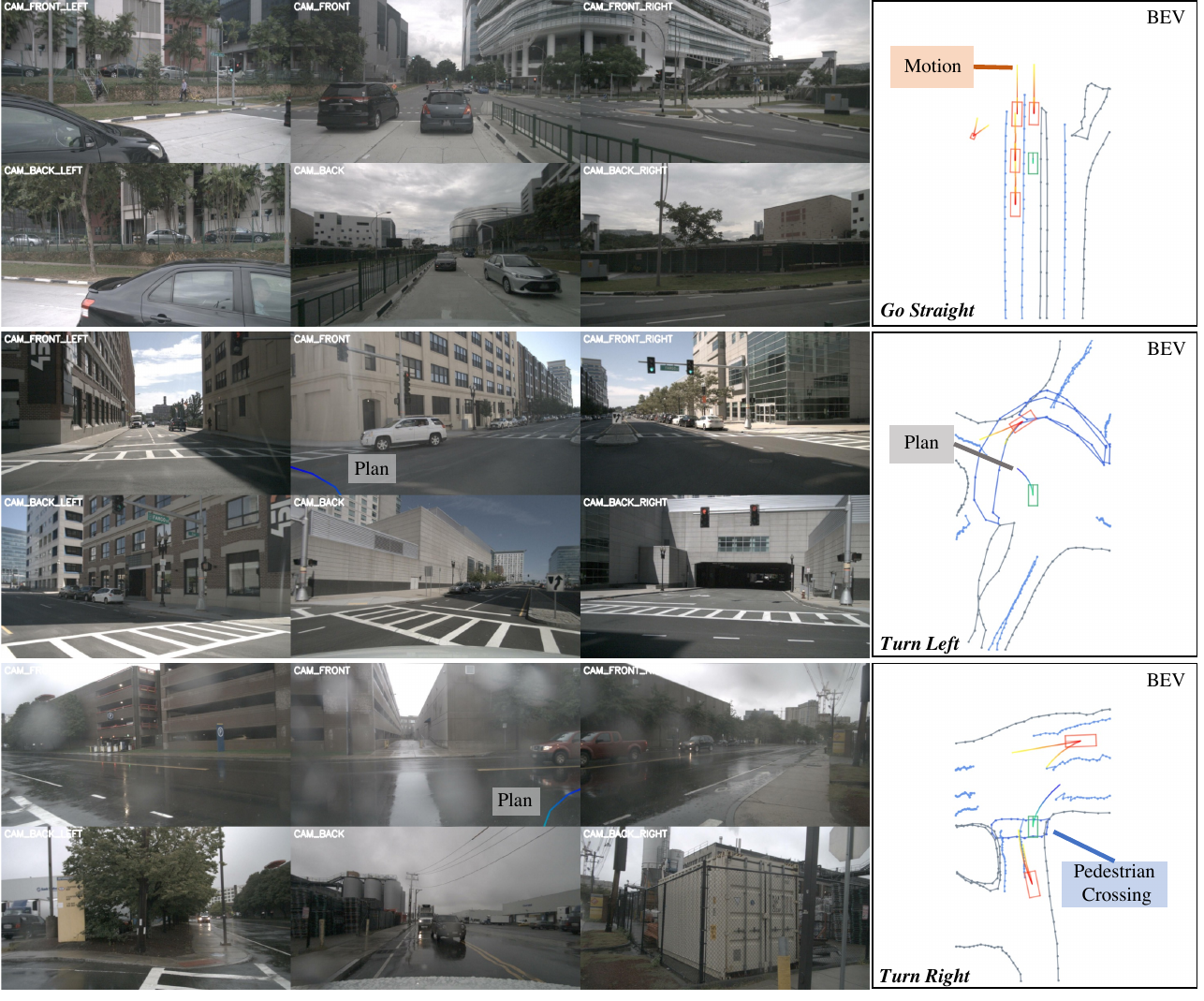}
 \vspace{-5pt}
	\caption{Qualitative results of EgoFSD. We omit the map selection results for clarity of road structure details.}
	\label{fig_visualization}
\vspace{-5pt}
\end{figure}

\section{Conclusion}


In this paper, we propose a \textbf{Fully Sparse Paradigm} for end-to-end self-driving in an \textbf{Ego-Centric} manner, termed as EgoFSD. EgoFSD conducts hierarchical interaction based on sparse representation and perception results. Only interactive agents are considered for joint motion prediction including the ego-vehicle. Iterative planning optimization strategy contributes to the driving safety with interactive decision. Besides, uncertainty modeling is conducted to improve the stability of end-to-end system. Extensive ablations and comparisons reveal the superiority and great potential of our ego-centric fully sparse paradigm.


{
    \small
    \bibliographystyle{ieeenat_fullname}
    \bibliography{main}

\begin{thebibliography}{56}
\providecommand{\natexlab}[1]{#1}
\providecommand{\url}[1]{\texttt{#1}}
\expandafter\ifx\csname urlstyle\endcsname\relax
  \providecommand{\doi}[1]{doi: #1}\else
  \providecommand{\doi}{doi: \begingroup \urlstyle{rm}\Url}\fi

\bibitem[Caesar et~al.(2020)Caesar, Bankiti, Lang, Vora, Liong, Xu, Krishnan, Pan, Baldan, and Beijbom]{caesar2020nuscenes}
Holger Caesar, Varun Bankiti, Alex~H Lang, Sourabh Vora, Venice~Erin Liong, Qiang Xu, Anush Krishnan, Yu Pan, Giancarlo Baldan, and Oscar Beijbom.
\newblock nuscenes: A multimodal dataset for autonomous driving.
\newblock In \emph{CVPR}, pages 11621--11631, 2020.

\bibitem[Casas et~al.(2018)Casas, Luo, and Urtasun]{casas2018intentnet}
Sergio Casas, Wenjie Luo, and Raquel Urtasun.
\newblock Intentnet: Learning to predict intention from raw sensor data.
\newblock In \emph{CoRL}, pages 947--956, 2018.

\bibitem[Codevilla et~al.(2018)Codevilla, M{\"u}ller, L{\'o}pez, Koltun, and Dosovitskiy]{codevilla2018end}
Felipe Codevilla, Matthias M{\"u}ller, Antonio L{\'o}pez, Vladlen Koltun, and Alexey Dosovitskiy.
\newblock End-to-end driving via conditional imitation learning.
\newblock In \emph{ICRA}, pages 4693--4700, 2018.

\bibitem[Codevilla et~al.(2019)Codevilla, Santana, L{\'o}pez, and Gaidon]{codevilla2019exploring}
Felipe Codevilla, Eder Santana, Antonio~M L{\'o}pez, and Adrien Gaidon.
\newblock Exploring the limitations of behavior cloning for autonomous driving.
\newblock In \emph{ICCV}, pages 9329--9338, 2019.

\bibitem[Dosovitskiy et~al.(2017)Dosovitskiy, Ros, Codevilla, Lopez, and Koltun]{dosovitskiy2017carla}
Alexey Dosovitskiy, German Ros, Felipe Codevilla, Antonio Lopez, and Vladlen Koltun.
\newblock Carla: An open urban driving simulator.
\newblock In \emph{Conference on Robot Learning}, pages 1--16, 2017.

\bibitem[Gu et~al.(2023)Gu, Hu, Zhang, Chen, Wang, Wang, and Zhao]{gu2023vip3d}
Junru Gu, Chenxu Hu, Tianyuan Zhang, Xuanyao Chen, Yilun Wang, Yue Wang, and Hang Zhao.
\newblock Vip3d: End-to-end visual trajectory prediction via 3d agent queries.
\newblock In \emph{Proceedings of the IEEE Conference on Computer Vision and Pattern Recognition}, pages 5496--5506, 2023.

\bibitem[Han et~al.(2024)Han, Yang, Sun, Ge, Dong, Zhou, Mao, Peng, and Zhang]{han2024exploring}
Chunrui Han, Jinrong Yang, Jianjian Sun, Zheng Ge, Runpei Dong, Hongyu Zhou, Weixin Mao, Yuang Peng, and Xiangyu Zhang.
\newblock Exploring recurrent long-term temporal fusion for multi-view 3d perception.
\newblock \emph{IEEE Robotics and Automation Letters}, 2024.

\bibitem[He et~al.(2016)He, Zhang, Ren, and Sun]{he2016deep}
Kaiming He, Xiangyu Zhang, Shaoqing Ren, and Jian Sun.
\newblock Deep residual learning for image recognition.
\newblock In \emph{Proceedings of the IEEE Conference on Computer Vision and Pattern Recognition}, pages 770--778, 2016.

\bibitem[Hu et~al.(2018)Hu, Shen, and Sun]{hu2018squeeze}
Jie Hu, Li Shen, and Gang Sun.
\newblock Squeeze-and-excitation networks.
\newblock In \emph{Proceedings of the IEEE Conference on Computer Vision and Pattern Recognition}, pages 7132--7141, 2018.

\bibitem[Hu et~al.(2021)Hu, Huang, Dolan, Held, and Ramanan]{hu2021safe}
Peiyun Hu, Aaron Huang, John Dolan, David Held, and Deva Ramanan.
\newblock Safe local motion planning with self-supervised freespace forecasting.
\newblock In \emph{Proceedings of the IEEE/CVF Conference on Computer Vision and Pattern Recognition}, pages 12732--12741, 2021.

\bibitem[Hu et~al.(2022)Hu, Chen, Wu, Li, Yan, and Tao]{hu2022st}
Shengchao Hu, Li Chen, Penghao Wu, Hongyang Li, Junchi Yan, and Dacheng Tao.
\newblock St-p3: End-to-end vision-based autonomous driving via spatial-temporal feature learning.
\newblock In \emph{ECCV}, pages 533--549, 2022.

\bibitem[Hu et~al.(2023)Hu, Yang, Chen, Li, Sima, Zhu, Chai, Du, Lin, Wang, et~al.]{hu2023planning}
Yihan Hu, Jiazhi Yang, Li Chen, Keyu Li, Chonghao Sima, Xizhou Zhu, Siqi Chai, Senyao Du, Tianwei Lin, Wenhai Wang, et~al.
\newblock Planning-oriented autonomous driving.
\newblock In \emph{Proceedings of the IEEE Conference on Computer Vision and Pattern Recognition}, pages 17853--17862, 2023.

\bibitem[Huang et~al.(2023)Huang, Li, Xie, Liang, Wang, Shen, Liu, Wang, Luo, and Shao]{huang2023fast}
Bin Huang, Yangguang Li, Enze Xie, Feng Liang, Luya Wang, Mingzhu Shen, Fenggang Liu, Tianqi Wang, Ping Luo, and Jing Shao.
\newblock Fast-bev: Towards real-time on-vehicle bird's-eye view perception.
\newblock \emph{arXiv preprint arXiv:2301.07870}, 2023.

\bibitem[Huang et~al.(2021)Huang, Huang, Zhu, Ye, and Du]{huang2021bevdet}
Junjie Huang, Guan Huang, Zheng Zhu, Yun Ye, and Dalong Du.
\newblock Bevdet: High-performance multi-camera 3d object detection in bird-eye-view.
\newblock \emph{arXiv preprint arXiv:2112.11790}, 2021.

\bibitem[Jia et~al.(2024)Jia, Yang, Li, Zhang, and Yan]{jia2024bench2drive}
Xiaosong Jia, Zhenjie Yang, Qifeng Li, Zhiyuan Zhang, and Junchi Yan.
\newblock Bench2drive: Towards multi-ability benchmarking of closed-loop end-to-end autonomous driving.
\newblock \emph{arXiv preprint arXiv:2406.03877}, 2024.

\bibitem[Jiang et~al.(2022)Jiang, Chen, Wang, Liao, Cheng, Chen, Zhou, Zhang, Liu, and Huang]{jiang2022perceive}
Bo Jiang, Shaoyu Chen, Xinggang Wang, Bencheng Liao, Tianheng Cheng, Jiajie Chen, Helong Zhou, Qian Zhang, Wenyu Liu, and Chang Huang.
\newblock Perceive, interact, predict: Learning dynamic and static clues for end-to-end motion prediction.
\newblock \emph{arXiv preprint arXiv:2212.02181}, 2022.

\bibitem[Jiang et~al.(2023)Jiang, Chen, Xu, Liao, Chen, Zhou, Zhang, Liu, Huang, and Wang]{jiang2023vad}
Bo Jiang, Shaoyu Chen, Qing Xu, Bencheng Liao, Jiajie Chen, Helong Zhou, Qian Zhang, Wenyu Liu, Chang Huang, and Xinggang Wang.
\newblock Vad: Vectorized scene representation for efficient autonomous driving.
\newblock In \emph{Proceedings of the IEEE International Conference on Computer Vision}, pages 8340--8350, 2023.

\bibitem[Jin et~al.(2025{\natexlab{a}})Jin, Su, Liu, Ma, Wu, Hui, and Yan]{jin2025unimamba}
Xin Jin, Haisheng Su, Kai Liu, Cong Ma, Wei Wu, Fei Hui, and Junchi Yan.
\newblock Unimamba: Unified spatial-channel representation learning with group-efficient mamba for lidar-based 3d object detection.
\newblock In \emph{Proceedings of the Computer Vision and Pattern Recognition Conference}, pages 1407--1417, 2025{\natexlab{a}}.

\bibitem[Jin et~al.(2025{\natexlab{b}})Jin, Su, Ma, Liu, Wu, Hui, and Yan]{jin2025geoformer}
Xin Jin, Haisheng Su, Cong Ma, Kai Liu, Wei Wu, Fei Hui, and Junchi Yan.
\newblock Geoformer: Geometry point encoder for 3d object detection with graph-based transformer.
\newblock In \emph{Proceedings of the IEEE/CVF International Conference on Computer Vision}, pages 26879--26889, 2025{\natexlab{b}}.

\bibitem[Khurana et~al.(2022)Khurana, Hu, Dave, Ziglar, Held, and Ramanan]{khurana2022differentiable}
Tarasha Khurana, Peiyun Hu, Achal Dave, Jason Ziglar, David Held, and Deva Ramanan.
\newblock Differentiable raycasting for self-supervised occupancy forecasting.
\newblock In \emph{ECCV}, pages 353--369, 2022.

\bibitem[Li et~al.(2023{\natexlab{a}})Li, Lyu, Ma, Wang, Yang, Li, and Li]{li2023normalization}
Lu Li, Jiafei Lyu, Guozheng Ma, Zilin Wang, Zhenjie Yang, Xiu Li, and Zhiheng Li.
\newblock Normalization enhances generalization in visual reinforcement learning.
\newblock \emph{arXiv preprint arXiv:2306.00656}, 2023{\natexlab{a}}.

\bibitem[Li et~al.(2022{\natexlab{a}})Li, Wang, Wang, and Zhao]{li2022hdmapnet}
Qi Li, Yue Wang, Yilun Wang, and Hang Zhao.
\newblock Hdmapnet: An online hd map construction and evaluation framework.
\newblock In \emph{2022 International Conference on Robotics and Automation (ICRA)}, pages 4628--4634, 2022{\natexlab{a}}.

\bibitem[Li et~al.(2023{\natexlab{b}})Li, Ge, Yu, Yang, Wang, Shi, Sun, and Li]{li2023bevdepth}
Yinhao Li, Zheng Ge, Guanyi Yu, Jinrong Yang, Zengran Wang, Yukang Shi, Jianjian Sun, and Zeming Li.
\newblock Bevdepth: Acquisition of reliable depth for multi-view 3d object detection.
\newblock In \emph{Proceedings of the AAAI Conference on Artificial Intelligence}, pages 1477--1485, 2023{\natexlab{b}}.

\bibitem[Li et~al.(2022{\natexlab{b}})Li, Wang, Li, Xie, Sima, Lu, Qiao, and Dai]{li2022bevformer}
Zhiqi Li, Wenhai Wang, Hongyang Li, Enze Xie, Chonghao Sima, Tong Lu, Yu Qiao, and Jifeng Dai.
\newblock Bevformer: Learning bird’s-eye-view representation from multi-camera images via spatiotemporal transformers.
\newblock In \emph{ECCV}, pages 1--18, 2022{\natexlab{b}}.

\bibitem[Li et~al.(2024)Li, Yu, Lan, Li, Kautz, Lu, and Alvarez]{li2024ego}
Zhiqi Li, Zhiding Yu, Shiyi Lan, Jiahan Li, Jan Kautz, Tong Lu, and Jose~M Alvarez.
\newblock Is ego status all you need for open-loop end-to-end autonomous driving?
\newblock In \emph{Proceedings of the IEEE Conference on Computer Vision and Pattern Recognition}, pages 14864--14873, 2024.

\bibitem[Liang et~al.(2020)Liang, Yang, Zeng, Chen, Hu, Casas, and Urtasun]{liang2020pnpnet}
Ming Liang, Bin Yang, Wenyuan Zeng, Yun Chen, Rui Hu, Sergio Casas, and Raquel Urtasun.
\newblock Pnpnet: End-to-end perception and prediction with tracking in the loop.
\newblock In \emph{Proceedings of the IEEE Conference on Computer Vision and Pattern Recognition}, pages 11553--11562, 2020.

\bibitem[Liao et~al.(2022)Liao, Chen, Wang, Cheng, Zhang, Liu, and Huang]{liao2022maptr}
Bencheng Liao, Shaoyu Chen, Xinggang Wang, Tianheng Cheng, Qian Zhang, Wenyu Liu, and Chang Huang.
\newblock Maptr: Structured modeling and learning for online vectorized hd map construction.
\newblock \emph{arXiv preprint arXiv:2208.14437}, 2022.

\bibitem[Liao et~al.(2024)Liao, Chen, Yin, Jiang, Wang, Yan, Zhang, Li, Zhang, Zhang, et~al.]{liao2024diffusiondrive}
Bencheng Liao, Shaoyu Chen, Haoran Yin, Bo Jiang, Cheng Wang, Sixu Yan, Xinbang Zhang, Xiangyu Li, Ying Zhang, Qian Zhang, et~al.
\newblock Diffusiondrive: Truncated diffusion model for end-to-end autonomous driving.
\newblock \emph{arXiv preprint arXiv:2411.15139}, 2024.

\bibitem[Lin et~al.(2023)Lin, Pei, Lin, Huang, and Su]{lin2023sparse4d}
Xuewu Lin, Zixiang Pei, Tianwei Lin, Lichao Huang, and Zhizhong Su.
\newblock Sparse4d v3: Advancing end-to-end 3d detection and tracking.
\newblock \emph{arXiv preprint arXiv:2311.11722}, 2023.

\bibitem[Liu et~al.(2023{\natexlab{a}})Liu, Teng, Lu, Wang, and Wang]{liu2023sparsebev}
Haisong Liu, Yao Teng, Tao Lu, Haiguang Wang, and Limin Wang.
\newblock Sparsebev: High-performance sparse 3d object detection from multi-camera videos.
\newblock In \emph{Proceedings of the IEEE International Conference on Computer Vision}, pages 18580--18590, 2023{\natexlab{a}}.

\bibitem[Liu et~al.(2022)Liu, Wang, Zhang, and Sun]{liu2022petr}
Yingfei Liu, Tiancai Wang, Xiangyu Zhang, and Jian Sun.
\newblock Petr: Position embedding transformation for multi-view 3d object detection.
\newblock In \emph{European Conference on Computer Vision}, pages 531--548, 2022.

\bibitem[Liu et~al.(2023{\natexlab{b}})Liu, Yan, Jia, Li, Gao, Wang, and Zhang]{liu2023petrv2}
Yingfei Liu, Junjie Yan, Fan Jia, Shuailin Li, Aqi Gao, Tiancai Wang, and Xiangyu Zhang.
\newblock Petrv2: A unified framework for 3d perception from multi-camera images.
\newblock In \emph{Proceedings of the IEEE International Conference on Computer Vision}, pages 3262--3272, 2023{\natexlab{b}}.

\bibitem[Liu et~al.(2023{\natexlab{c}})Liu, Yuan, Wang, Wang, and Zhao]{liu2023vectormapnet}
Yicheng Liu, Tianyuan Yuan, Yue Wang, Yilun Wang, and Hang Zhao.
\newblock Vectormapnet: End-to-end vectorized hd map learning.
\newblock In \emph{International Conference on Machine Learning}, pages 22352--22369, 2023{\natexlab{c}}.

\bibitem[Loshchilov and Hutter(2016)]{loshchilov2016sgdr}
Ilya Loshchilov and Frank Hutter.
\newblock Sgdr: Stochastic gradient descent with warm restarts.
\newblock \emph{arXiv preprint arXiv:1608.03983}, 2016.

\bibitem[Loshchilov and Hutter(2017)]{loshchilov2017decoupled}
Ilya Loshchilov and Frank Hutter.
\newblock Decoupled weight decay regularization.
\newblock \emph{arXiv preprint arXiv:1711.05101}, 2017.

\bibitem[Luo et~al.(2018)Luo, Yang, and Urtasun]{luo2018fast}
Wenjie Luo, Bin Yang, and Raquel Urtasun.
\newblock Fast and furious: Real time end-to-end 3d detection, tracking and motion forecasting with a single convolutional net.
\newblock In \emph{Proceedings of the IEEE Conference on Computer Vision and Pattern Recognition}, pages 3569--3577, 2018.

\bibitem[Niu et~al.(2023)Niu, Pu, Yang, Li, Zhou, Ren, Hu, Li, and Liu]{niu2023lightzero}
Yazhe Niu, Yuan Pu, Zhenjie Yang, Xueyan Li, Tong Zhou, Jiyuan Ren, Shuai Hu, Hongsheng Li, and Yu Liu.
\newblock Lightzero: A unified benchmark for monte carlo tree search in general sequential decision scenarios.
\newblock \emph{Advances in Neural Information Processing Systems}, 36:\penalty0 37594--37635, 2023.

\bibitem[Philion and Fidler(2020)]{philion2020lift}
Jonah Philion and Sanja Fidler.
\newblock Lift, splat, shoot: Encoding images from arbitrary camera rigs by implicitly unprojecting to 3d.
\newblock In \emph{Computer Vision--ECCV 2020: 16th European Conference, Glasgow, UK, August 23--28, 2020, Proceedings, Part XIV 16}, pages 194--210. Springer, 2020.

\bibitem[Pu et~al.(2024)Pu, Niu, Yang, Ren, Li, and Liu]{pu2024unizero}
Yuan Pu, Yazhe Niu, Zhenjie Yang, Jiyuan Ren, Hongsheng Li, and Yu Liu.
\newblock Unizero: Generalized and efficient planning with scalable latent world models.
\newblock \emph{arXiv preprint arXiv:2406.10667}, 2024.

\bibitem[Song et~al.(2025)Song, Jia, Liu, Pan, Zhang, Wang, Zhang, Xu, Yang, and Luo]{song2025don}
Ziying Song, Caiyan Jia, Lin Liu, Hongyu Pan, Yongchang Zhang, Junming Wang, Xingyu Zhang, Shaoqing Xu, Lei Yang, and Yadan Luo.
\newblock Don't shake the wheel: Momentum-aware planning in end-to-end autonomous driving.
\newblock \emph{arXiv preprint arXiv:2503.03125}, 2025.

\bibitem[Su et~al.(2024)Su, Song, Ma, Wu, and Yan]{su2024robosense}
Haisheng Su, Feixiang Song, Cong Ma, Wei Wu, and Junchi Yan.
\newblock Robosense: Large-scale dataset and benchmark for egocentric robot perception and navigation in crowded and unstructured environments.
\newblock \emph{arXiv preprint arXiv:2408.15503}, 2024.

\bibitem[Su et~al.(2025)Su, Zhang, Song, Zhou, Wu, Yan, and Zheng]{su2025freqpde}
Haisheng Su, Junjie Zhang, Feixiang Song, Sanping Zhou, Wei Wu, Junchi Yan, and Nanning Zheng.
\newblock Freqpde: Rethinking positional depth embedding for multi-view 3d object detection transformers.
\newblock In \emph{Proceedings of the IEEE/CVF International Conference on Computer Vision}, pages 28145--28155, 2025.

\bibitem[Sun et~al.(2024)Sun, Lin, Shi, Zhang, Wu, and Zheng]{sun2024sparsedrive}
Wenchao Sun, Xuewu Lin, Yining Shi, Chuang Zhang, Haoran Wu, and Sifa Zheng.
\newblock Sparsedrive: End-to-end autonomous driving via sparse scene representation.
\newblock \emph{arXiv preprint arXiv:2405.19620}, 2024.

\bibitem[Wang et~al.(2023)Wang, Liu, Wang, Li, and Zhang]{wang2023exploring}
Shihao Wang, Yingfei Liu, Tiancai Wang, Ying Li, and Xiangyu Zhang.
\newblock Exploring object-centric temporal modeling for efficient multi-view 3d object detection.
\newblock In \emph{ICCV}, pages 3621--3631, 2023.

\bibitem[Wang et~al.(2025)Wang, Zhang, Qu, Li, Liu, and Huang]{wang2025diffad}
Tao Wang, Cong Zhang, Xingguang Qu, Kun Li, Weiwei Liu, and Chang Huang.
\newblock Diffad: A unified diffusion modeling approach for autonomous driving.
\newblock \emph{arXiv preprint arXiv:2503.12170}, 2025.

\bibitem[Wang et~al.(2022)Wang, Guizilini, Zhang, Wang, Zhao, and Solomon]{wang2022detr3d}
Yue Wang, Vitor~Campagnolo Guizilini, Tianyuan Zhang, Yilun Wang, Hang Zhao, and Justin Solomon.
\newblock Detr3d: 3d object detection from multi-view images via 3d-to-2d queries.
\newblock In \emph{Conference on Robot Learning}, pages 180--191, 2022.

\bibitem[Yang et~al.(2023{\natexlab{a}})Yang, Chen, Tian, Tao, Zhu, Zhang, Huang, Li, Qiao, Lu, et~al.]{yang2023bevformer}
Chenyu Yang, Yuntao Chen, Hao Tian, Chenxin Tao, Xizhou Zhu, Zhaoxiang Zhang, Gao Huang, Hongyang Li, Yu Qiao, Lewei Lu, et~al.
\newblock Bevformer v2: Adapting modern image backbones to bird's-eye-view recognition via perspective supervision.
\newblock In \emph{CVPR}, pages 17830--17839, 2023{\natexlab{a}}.

\bibitem[Yang et~al.(2023{\natexlab{b}})Yang, Jia, Li, and Yan]{yang2023llm4drive}
Zhenjie Yang, Xiaosong Jia, Hongyang Li, and Junchi Yan.
\newblock Llm4drive: A survey of large language models for autonomous driving.
\newblock \emph{arXiv preprint arXiv:2311.01043}, 2023{\natexlab{b}}.

\bibitem[Yang et~al.(2025{\natexlab{a}})Yang, Chai, Jia, Li, Shao, Zhu, Su, and Yan]{yang2025drivemoe}
Zhenjie Yang, Yilin Chai, Xiaosong Jia, Qifeng Li, Yuqian Shao, Xuekai Zhu, Haisheng Su, and Junchi Yan.
\newblock Drivemoe: Mixture-of-experts for vision-language-action model in end-to-end autonomous driving.
\newblock \emph{arXiv preprint arXiv:2505.16278}, 2025{\natexlab{a}}.

\bibitem[Yang et~al.(2025{\natexlab{b}})Yang, Jia, Li, Yang, Yao, and Yan]{yang2025raw2drive}
Zhenjie Yang, Xiaosong Jia, Qifeng Li, Xue Yang, Maoqing Yao, and Junchi Yan.
\newblock Raw2drive: Reinforcement learning with aligned world models for end-to-end autonomous driving (in carla v2).
\newblock \emph{arXiv preprint arXiv:2505.16394}, 2025{\natexlab{b}}.

\bibitem[Ye et~al.(2023)Ye, Jing, Hu, Huang, Gao, Li, Wang, Guo, Xiao, Mao, et~al.]{ye2023fusionad}
Tengju Ye, Wei Jing, Chunyong Hu, Shikun Huang, Lingping Gao, Fangzhen Li, Jingke Wang, Ke Guo, Wencong Xiao, Weibo Mao, et~al.
\newblock Fusionad: Multi-modality fusion for prediction and planning tasks of autonomous driving.
\newblock \emph{arXiv preprint arXiv:2308.01006}, 2023.

\bibitem[Yuan et~al.(2024)Yuan, Liu, Wang, Wang, and Zhao]{yuan2024streammapnet}
Tianyuan Yuan, Yicheng Liu, Yue Wang, Yilun Wang, and Hang Zhao.
\newblock Streammapnet: Streaming mapping network for vectorized online hd map construction.
\newblock In \emph{Proceedings of the IEEE Winter Conference on Applications of Computer Vision}, pages 7356--7365, 2024.

\bibitem[Zhai et~al.(2023)Zhai, Feng, Du, Mao, Liu, Tan, Zhang, Ye, and Wang]{zhai2023rethinking}
Jiang-Tian Zhai, Ze Feng, Jinhao Du, Yongqiang Mao, Jiang-Jiang Liu, Zichang Tan, Yifu Zhang, Xiaoqing Ye, and Jingdong Wang.
\newblock Rethinking the open-loop evaluation of end-to-end autonomous driving in nuscenes.
\newblock \emph{arXiv preprint arXiv:2305.10430}, 2023.

\bibitem[Zhang et~al.(2024{\natexlab{a}})Zhang, Wang, Zhu, Zhao, Chen, Zhang, Gong, Zhou, Zhang, Wang, et~al.]{zhang2024sparsead}
Diankun Zhang, Guoan Wang, Runwen Zhu, Jianbo Zhao, Xiwu Chen, Siyu Zhang, Jiahao Gong, Qibin Zhou, Wenyuan Zhang, Ningzi Wang, et~al.
\newblock Sparsead: Sparse query-centric paradigm for efficient end-to-end autonomous driving.
\newblock \emph{arXiv preprint arXiv:2404.06892}, 2024{\natexlab{a}}.

\bibitem[Zhang et~al.(2024{\natexlab{b}})Zhang, Qian, Li, Pan, Chen, Liang, Zhang, Zhang, Li, Fu, et~al.]{zhang2024graphad}
Yunpeng Zhang, Deheng Qian, Ding Li, Yifeng Pan, Yong Chen, Zhenbao Liang, Zhiyao Zhang, Shurui Zhang, Hongxu Li, Maolei Fu, et~al.
\newblock Graphad: Interaction scene graph for end-to-end autonomous driving.
\newblock \emph{arXiv preprint arXiv:2403.19098}, 2024{\natexlab{b}}.

\bibitem[Zheng et~al.(2024)Zheng, Song, Guo, Zhang, and Chen]{zheng2024genad}
Wenzhao Zheng, Ruiqi Song, Xianda Guo, Chenming Zhang, and Long Chen.
\newblock Genad: Generative end-to-end autonomous driving.
\newblock \emph{arXiv preprint arXiv: 2402.11502}, 2024.

\end{thebibliography}
}

\clearpage
\maketitlesupplementary
\appendix
\setcounter{page}{1}
\setcounter{section}{0}
\setcounter{figure}{0}
\setcounter{table}{0}
\setcounter{equation}{0}
\renewcommand{\thefigure}{A\arabic{figure}}

\section{Evaluation Metrics}
\label{app:metric}

\noindent
\textbf{Perception.} The evaluation for detection and online mapping follows standard evaluation protocols~\cite{caesar2020nuscenes}. For detection, we use mean Average Precision (mAP), mean Average Error of Translation (mATE), Scale (mASE), Orientation (mAOE), Velocity (mAVE), Attribute (mAAE) and nuScenes Detection Score (NDS) to evaluate the model performance. For online mapping, we calculate the Average Precision (AP) of three map classes: lane divider, pedestrian crossing and road boundary, then average across all classes to get mean Average Precision (mAP).

\noindent
\textbf{Open-Loop Planning.} We adopt commonly used L2 error and collision rate to evaluate the planning performance. The evaluation of L2 error is aligned with VAD~\cite{jiang2023vad}. For collision rate, there are two drawbacks in previous ~\cite{hu2023planning,jiang2023vad} implementation, resulting in inaccurate evaluation in planning performance. On one hand, previous benchmark convert obstacle bounding boxes into occupancy map with a grid size of 0.5m, resulting in false collisions in certain cases, e.g. ego vehicle approaches obstacles that smaller than a single occupancy map pixel~\cite{zhai2023rethinking}. (2) The heading of ego vehicle is not considered and assumed to remain unchanged~\cite{li2024ego}. To accurately evaluate the planning performance, we account for the changes in ego heading by estimating the yaw angle through trajectory points, and assess the presence of a collision by examining the overlap between the bounding boxes of ego vehicle and obstacles. We reproduce the planning results on our benchmark with official checkpoints~\cite{hu2023planning,jiang2023vad} for a fair comparison.

\noindent
\textbf{Closed-Loop Planning.} We use the official 220 routes and official metrics of Bench2Drive~\cite{jia2024bench2drive} for evaluation. The Driving Score (DS) is defined as the product of Route Completion and Infraction Score, capturing both task completion and rule adherence. The Success Rate (SR) measures the percentage of routes completed successfully within the allocated time. A route is deemed successful if the ego vehicle reaches its destination without any traffic rule violation. Efficiency quantifies the vehicle’s velocity relative to surrounding traffic, encouraging progressiveness without aggression.

\section{More Details}

\noindent
\textbf{Problem Formulation.} Given multi-view camera image sequence can be denoted as $\mathbf{S}=\{I_{t} \in \mathbb{R}^{N\times 3\times H \times W}\}_{t=T-k}^{T}$, where $N$ is the number of camera views and $k$ indicates the temporal length till current timestep $T$ respectively. Annotation of input $\mathbf{S}$ for end-to-end planning is composed by a set of future waypoints of the ego-vehicle $\psi = \{\phi=(x_t, y_t)\}_{t=1}^{T_{e}}$, where $T_{e}$ = 3 seconds (2Hz) is the planning time horizon, and ($x_t$, $y_t$) is the BEV location transformed to the ego-vehicle coordinate system at current timestep $T$. Meanwhile, driving command as well as ego-status is also provided. Annotation set $\psi$ is used during training. During prediction, the planned trajectory of ego-vehicle should fit the annotation $\psi$ with minimum L2 errors and collision rate with surrounding agents.

\noindent
\textbf{Data Augmentation.} During training phase, we conduct a global data augmentation strategy to improve the model stability. Specifically, random rotation, translation and flipping (Y-axis) are applied to both bounding boxes of agent / map instances and future trajectories of agents / ego-vehicle respectively. Meanwhile, we adjust the camera extrinsic parameters accordingly to ensure spatial alignment between the surround-view images and the noise-perturbed targets.

\section{More Ablation Study}
\label{app:ablation}

\noindent
\textbf{Necessity and Order of Object Selection.} Tab.~\ref{table_selector} studies the necessity of agent and map selection during the ego-centric hierarchical interaction. We can observe that agent selection contributes more than the map selection, especially in the driving safety. And both of agent and map interactions are conducted in a cascaded order is inferior than the parallel manner, where the updated ego query from parallel outputs are concatenated for joint motion prediction.

\noindent
\textbf{Effect of Interactive Score Fusion.} During the ego-centric query selection, both geometric and classification scores are considered to ensure that the selected closest in-path queries are true positive agents or maps, which are adopted for motion planner. Tab.~\ref{tab_score} shows the effect of three types of scores used for query ranking, namely attention, geometry and confidence scores. As Eq.~\ref{eq1}, interactive score $S_{inter}$ obtained by multiplying these three scores can achieve the best selection quality and planning performance. $S_{inter}$ without confidence score fails to distinguish between background and foreground queries, resulting in inferior performance.

\begin{table}[tb!] \footnotesize
\begin{center}
\caption{Necessity of \textbf{Agent/Map Selection} and effect of \textbf{Interaction Order} in the hierarchical interaction module.}
\label{table_selector}
\vspace{-0.3cm}
\resizebox{1.\columnwidth}{!}{
\begin{tabular}{cc|cc|cccc|cccc}
\toprule[1pt]
\multirow{2}{*}{\makecell{\textbf{Agent} \\ \textbf{Selection}}} & \multirow{2}{*}{\makecell{\textbf{Map}\\ \textbf{Selection}}} & \multirow{2}{*}{\textbf{Cascade}} & \multirow{2}{*}{\textbf{Parallel}} & \multicolumn{4}{c|}{\textbf{Planning L2 (\textit{m}) $\downarrow$}} & \multicolumn{4}{c}{\textbf{Planning Coll. (\%)} $\downarrow$} \\ 
 
 & & & & \multicolumn{1}{c}{1\textit{s}} & \multicolumn{1}{c}{2\textit{s}} & \multicolumn{1}{c}{3\textit{s}} & \cellcolor[HTML]{DADADA}Avg. & \multicolumn{1}{c}{1\textit{s}} & \multicolumn{1}{c}{2\textit{s}} & \multicolumn{1}{c}{3\textit{s}} & \cellcolor[HTML]{DADADA}Avg. \\ \midrule
 
 
 \checkmark & \ding{55} & - & - & \multicolumn{1}{c}{0.16} & \multicolumn{1}{c}{0.34} & \multicolumn{1}{c}{0.64} & \cellcolor[HTML]{DADADA}0.38 & \multicolumn{1}{c}{0.03} & \multicolumn{1}{c}{0.05} & \multicolumn{1}{c}{0.22} & \cellcolor[HTML]{DADADA}0.10 \\ 
 
 \ding{55} & \checkmark & - & - & \multicolumn{1}{c}{0.17} & \multicolumn{1}{c}{0.35} & \multicolumn{1}{c}{0.63} & \cellcolor[HTML]{DADADA}0.38 & \multicolumn{1}{c}{0.02} & \multicolumn{1}{c}{0.06} & \multicolumn{1}{c}{0.28} & \cellcolor[HTML]{DADADA}0.12 \\ 

 \checkmark & \checkmark & \checkmark & - & \multicolumn{1}{c}{0.16} & \multicolumn{1}{c}{0.34} & \multicolumn{1}{c}{0.62} & \cellcolor[HTML]{DADADA}0.37 & \multicolumn{1}{c}{0.05} & \multicolumn{1}{c}{0.07} & \multicolumn{1}{c}{0.30} & \cellcolor[HTML]{DADADA}0.14 \\ 

 \checkmark & \checkmark & - & \checkmark & \multicolumn{1}{c}{0.16} & \multicolumn{1}{c}{0.33} & \multicolumn{1}{c}{0.59} & \cellcolor[HTML]{DADADA}\textbf{0.35} & \multicolumn{1}{c}{0.00} & \multicolumn{1}{c}{0.04} & \multicolumn{1}{c}{0.18} & \cellcolor[HTML]{DADADA}\textbf{0.07} \\ 
 \bottomrule[1pt]
\end{tabular}}
\end{center}
\end{table}

\begin{table}[tb!]\footnotesize
\begin{center}
\caption{Effect of \textbf{Interactive Score Fusion} process in the geometric attended selection step.} \label{tab_score}
\vspace{-0.3cm}
\resizebox{1.0\columnwidth}{!}{
\begin{tabular}{ccc|cccc|cccc}
\toprule[1pt]
\multirow{2}{*}{\makecell{\textbf{Attention}\\ \textbf{Score}}} & \multirow{2}{*}{\makecell{\textbf{Geometric} \\ \textbf{Score}}} & \multirow{2}{*}{\makecell{\textbf{Classification} \\ \textbf{Score}}} & \multicolumn{4}{c|}{\textbf{Planning L2 (\textit{m}) $\downarrow$}} & \multicolumn{4}{c}{\textbf{Planning Coll. (\%)} $\downarrow$} \\ 
 & &  & \multicolumn{1}{c}{1\textit{s}} & \multicolumn{1}{c}{2\textit{s}} & \multicolumn{1}{c}{3\textit{s}} & \cellcolor[HTML]{DADADA}Avg. & \multicolumn{1}{c}{1\textit{s}} & \multicolumn{1}{c}{2\textit{s}} & \multicolumn{1}{c}{3\textit{s}} & \cellcolor[HTML]{DADADA}Avg. \\ \midrule
 
   \checkmark & \ding{55} & \ding{55} & \multicolumn{1}{c}{0.18} & \multicolumn{1}{c}{0.36} & \multicolumn{1}{c}{0.66} & \cellcolor[HTML]{DADADA}0.39 & \multicolumn{1}{c}{0.09} & \multicolumn{1}{c}{0.11} & \multicolumn{1}{c}{0.28} & \cellcolor[HTML]{DADADA}0.16 \\ 

  \checkmark &  \checkmark & \ding{55} & \multicolumn{1}{c}{0.17} & \multicolumn{1}{c}{0.35} & \multicolumn{1}{c}{0.65} & \cellcolor[HTML]{DADADA}0.38 & \multicolumn{1}{c}{0.01} & \multicolumn{1}{c}{0.07} & \multicolumn{1}{c}{0.24} & \cellcolor[HTML]{DADADA}0.11 \\

  \checkmark & \checkmark & \checkmark & \multicolumn{1}{c}{0.16} & \multicolumn{1}{c}{0.33} & \multicolumn{1}{c}{0.59} & \cellcolor[HTML]{DADADA}\textbf{0.35} & \multicolumn{1}{c}{0.00} & \multicolumn{1}{c}{0.04} & \multicolumn{1}{c}{0.18} & \cellcolor[HTML]{DADADA}\textbf{0.07} \\
 \bottomrule[1pt]
\end{tabular}}
\end{center}
\end{table}

\noindent
\textbf{Effect of Sparse Perception.} Recent end-to-end planning method~\cite{sun2024sparsedrive} resorts to the sparse perception fashion to provide advanced 3D detection and online mapping results with high efficiency. To study the significance of advanced perception encoders for ego-planning, we compare the perception performance of various end-to-end methods as shown in Tab.~\ref{tab_perception}. With sparse perception encoder~\cite{lin2023sparse4d}, the performance of 3D object detection and online mapping can be greatly improved (10.6 NDS and 7.5 mAP, respectively) compared with dense BEV-based perception paradigm~\cite{li2022bevformer}. And the end-to-end planner~\cite{sun2024sparsedrive} equipped with the advanced perception encoder can consistently boost the planning performance. Therefore, the perception performance is essential for the end-to-end planner.



\begin{table*}[t]
\begin{center}
\caption{\textbf{Comparison of Perception Results on nuScenes val Dataset}. $\dag$: Reproduced with official checkpoint. $*$ indicates to use pre-trained weights from the nuImage dataset.} 
\label{tab_perception}
\vspace{-0.3cm}
    \centering
    \begin{subtable}{.45\textwidth}
        \centering
        \resizebox{1.0\columnwidth}{!}{
        \begin{tabular}{l|c|c|c|c}
        \toprule[1pt]
        \textbf{Method} & \textbf{Backbone} & \textbf{BEV} & \textbf{mAP} $\uparrow$ & \cellcolor[HTML]{DADADA}\textbf{NDS} $\uparrow$ \\ 
        \midrule 
         BEVFormer~\cite{li2022bevformer} &  ResNet101-DCN & \checkmark & 41.6 & \cellcolor[HTML]{DADADA}51.7 \\ 
         Sparse4Dv3~\cite{lin2023sparse4d} &  ResNet101$^{*}$ & \ding{55} & 53.7 & \cellcolor[HTML]{DADADA}62.3 \\ 
         \midrule
         UniAD~\cite{hu2023planning} &  ResNet101-DCN & \checkmark & 38.0 & \cellcolor[HTML]{DADADA}49.8 \\ 

         VAD-Base$^{\dag}$~\cite{jiang2023vad}  & ResNet50  &  \checkmark & 31.3 & \cellcolor[HTML]{DADADA}43.6 \\ 
        
         SparseDrive-S~\cite{sun2024sparsedrive} & ResNet50 & \ding{55} & 41.8 & \cellcolor[HTML]{DADADA}52.5  \\ 
        
         SparseDrive-B~\cite{sun2024sparsedrive} & ResNet101*  & \ding{55} & 49.6 & \cellcolor[HTML]{DADADA}58.8 \\ \midrule
         
         EgoFSD-S (BEV) & ResNet50 & \checkmark & 32.8 & \cellcolor[HTML]{DADADA}45.8 \\ 

         EgoFSD-S (PV) & ResNet50 & \ding{55} & 41.0 & \cellcolor[HTML]{DADADA}52.8 \\ 

         EgoFSD-B (PV) & ResNet101* & \ding{55} & 49.6 & \cellcolor[HTML]{DADADA}58.9 \\ 
         \bottomrule[1pt]
        \end{tabular}}
        \caption{3D detection results.}
    \end{subtable}%
    \begin{subtable}{.53\textwidth}
        \centering
        \resizebox{0.95\columnwidth}{!}{
        \begin{tabular}{l|ccc|c}
        \toprule[1pt]
        \textbf{Method} & \textbf{AP$_{ped}$} $\uparrow$ & \textbf{AP$_{divider}$} $\uparrow$ & \textbf{AP$_{boundary}$} $\uparrow$ & \cellcolor[HTML]{DADADA}\textbf{mAP} $\uparrow$ \\ 
        \midrule 
         VectorMapNet~\cite{liu2023vectormapnet} & 36.1 & 47.3 & 39.3 & \cellcolor[HTML]{DADADA}40.9  \\ 
         
         MapTR~\cite{liao2022maptr} & 56.2 & 59.8 & 60.1 & \cellcolor[HTML]{DADADA}58.7  \\ 
         \midrule
         
         VAD-Base$^{\dag}$~\cite{jiang2023vad} & 42.5 & 50.5 & 49.8 & \cellcolor[HTML]{DADADA}47.6   \\ 

        
         SparseDrive-S~\cite{sun2024sparsedrive}  & 49.9 & 57.0 & 58.4 & \cellcolor[HTML]{DADADA}55.1  \\ 
        
         SparseDrive-B~\cite{sun2024sparsedrive}  & 53.2 & 56.3 & 59.1 & \cellcolor[HTML]{DADADA}56.2  \\ \midrule
        
         EgoFSD-S (BEV)  & 46.7 & 54.3 & 56.0 & \cellcolor[HTML]{DADADA}52.3   \\ 
         EgoFSD-S (PV) & 54.9 & 55.7 & 57.3 & \cellcolor[HTML]{DADADA}56.0  \\ 
         EgoFSD-B (PV) & 52.3 & 58.2 & 59.3 & \cellcolor[HTML]{DADADA}56.6   \\ 
         \bottomrule[1pt]
        \end{tabular}}
        \caption{Online mapping results.}
    \end{subtable}
\end{center}
\end{table*}

\begin{figure}[tb!]
	\centering
	\includegraphics[width=8cm]{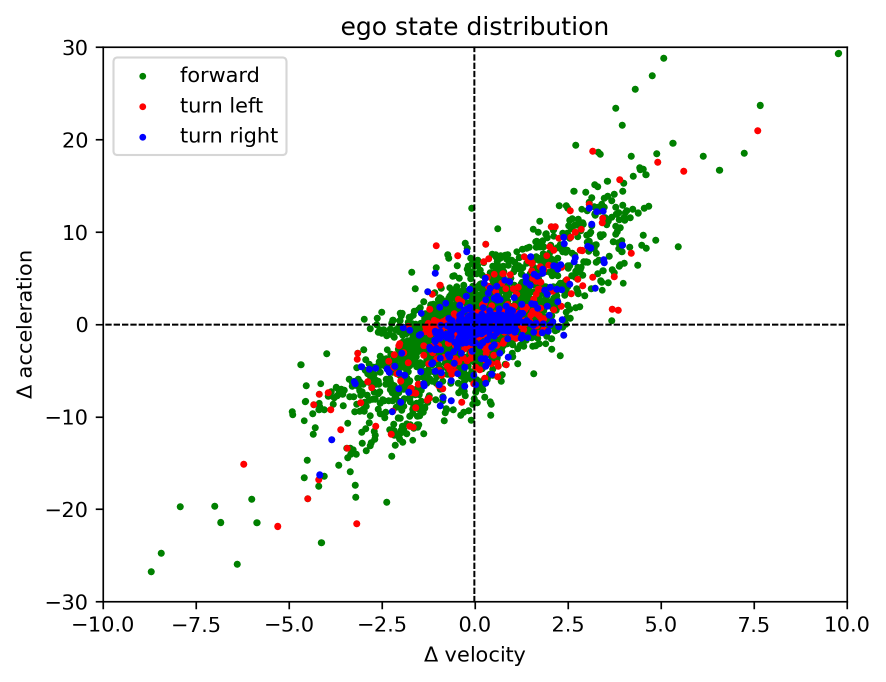}
    \vspace{-5pt}
	\caption{Distribution of Ground-Truth future ego states ($\Delta v$ vs. $\Delta a$) with different driving commands on the nuScenes \textbf{val} set.}
	\label{fig_state_distribution}
    \vspace{-5pt}
\end{figure}

\section{Effectiveness Analysis}
\begin{figure}[tb!]
    \vspace{-5pt}
	\centering
	\includegraphics[width=0.9\linewidth]{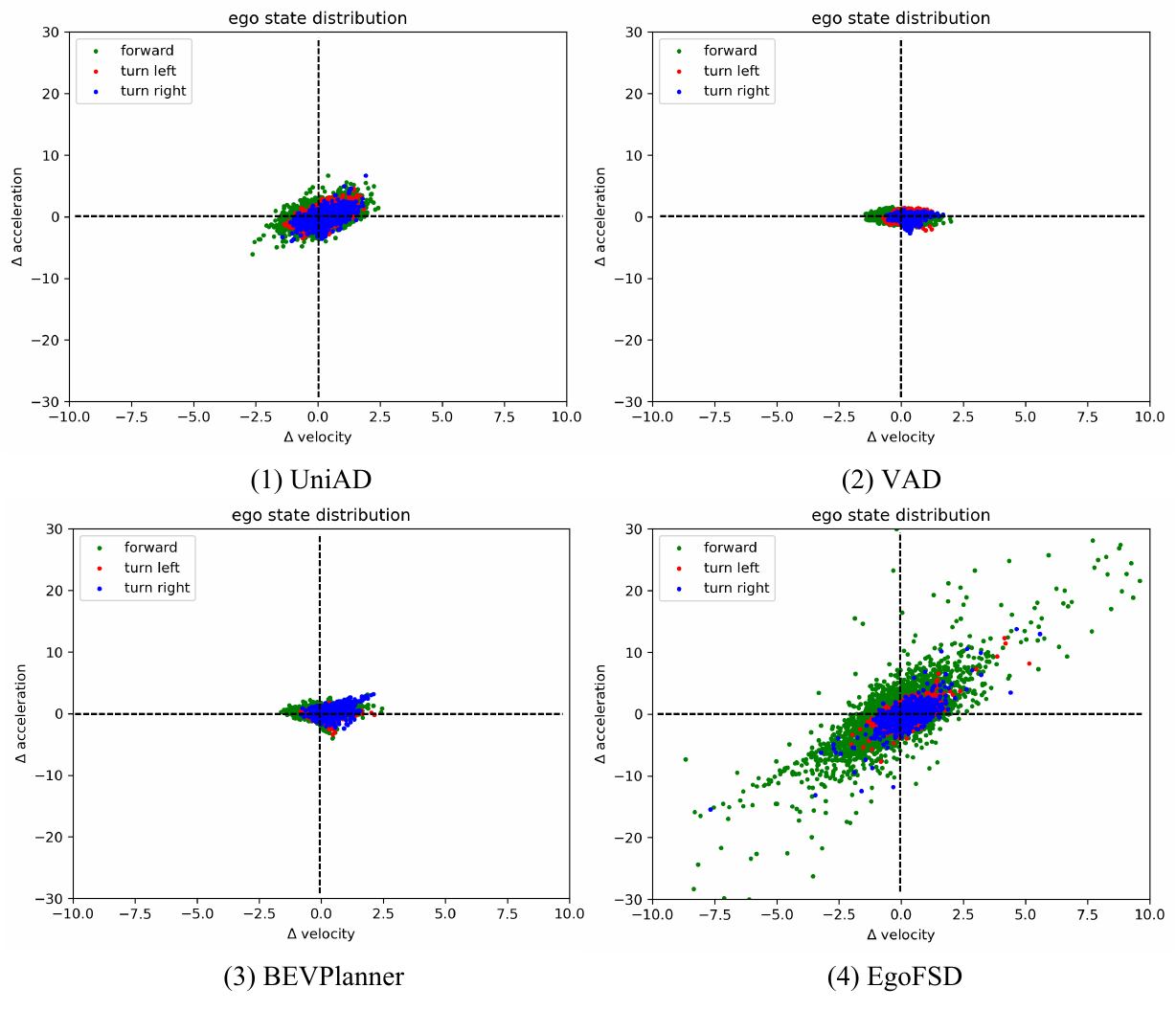}
	\caption{Comparison of predicted future ego state distribution of different end-to-end methods calculated by transforming predicted future waypoints on the nuScenes \textit{val} set.}
	\label{fig_state_dis_comparison}
\vspace{-5pt}
\end{figure}

The Ground-Truth future state distribution of ego-vehicle on nuScenes validation set is illustrated in Fig.~\ref{fig_state_distribution}, which is calculated with fixed time interval (1 second) between consecutive output future waypoints. And we also compare the output ego-state distribution of different popular E2E methods based on planned trajectories respectively, as show in Fig.~\ref{fig_state_dis_comparison}. We can observe that without ego-centric design, \textbf{existing end-to-end models are incapable of handling various emergencies appearing in the driving scenarios, where the absolute values of $\Delta v$ and $\Delta a$ are larger than normal situations (\textit{i.e.,} emergent braking, acceleration, turning or obstacle avoiance)}. Under this circumstance, the output planned trajectories cannot conform to the recorded expert demonstrations as expected. However, our EgoFSD performs consistently better in planning the future trajectory with dynamic and adaptive ego states of variable speed and acceleration. \textbf{Owing to the ego-centric hierarchical interaction and selection mechanism, the iterative motion planner can focus on the interactive agents rather than general objects}.

\section{More Discussion}
\label{app:discussion}

\noindent
\textbf{Practicality and Generalizability of Ego-Centric Design.} We empirically validate the necessity and effectiveness of our proposed ego-centric query selection and interactive planning with the help of geometric prior in Sec.~\ref{sec:ablation}. Moreover, we claim that the usage of geometric prior through interactive score fusion is \textbf{general and practical} for data-driven E2E-AD training, which considers both semantic relevance of driving context and human-like driving attention distribution according to current driving situation and future goal waypoint provided by the route planner. Extensive experiments conducted on both popular nuScenes~\cite{caesar2020nuscenes} and large-scale challenging Bench2Drive~\cite{jia2024bench2drive} benchmark demonstrate the generalizability and significant improvement of computational efficiency.



\noindent
\textbf{Comparison to Diffusion-based Planners.} We notice that recent works~\cite{liao2024diffusiondrive,wang2025diffad} also explore diffusion model for end-to-end autonomous driving application. DiffusionDrive~\cite{liao2024diffusiondrive} learns denoising from anchored Gaussian distribution to the multi-mode driving action distribution with a truncated diffusion policy. DiffAD~\cite{wang2025diffad} treats E2E-AD as a conditional image generation task, which adopts diffusion model to predict noise from the noisy latent BEV image. \textbf{Differently, we conduct a two-level data augmentation strategy to enrich the training samples including position diffusion of detected agents and trajectory diffusion of proposal ego trajectory}. In this way,  the joint motion prediction result can be more robust even with uncertain agent positions. And the offset prediction of ego trajectory can be more accurate in the planning optimization stage. Extensive ablations demonstrate the effectiveness of our uncertainty denoising design in improving the performance and training stability of EgoFSD.


\noindent
\textbf{Comparison to other Sparse Frameworks.} Prior to our work, some exiting methods~\cite{sun2024sparsedrive, zhang2024sparsead} are also built upon sparse representation for end-to-end autonomous driving. SparseDrive~\cite{sun2024sparsedrive} introduces the symmetric sparse perception architecture for object detection and online mapping respectively. However, it consumes more computation cost due to the repeated query projection, deformable feature aggregation, without a shared BEV feature as~\cite{jiang2023vad, hu2023planning}. Our EgoFSD constructs a shared 3D position-aware feature as~\cite{han2024exploring} to facilitate parallel sparse query decoding for perception and planning. Besides, an additional core contribution of our EgoFSD lies on an ego-centric design, which aims to select the most important agent/map instances (\textit{i.e.,} CIPV/CIPS) for interactive modeling. Benefiting from the geometric prior and iterative refinement, the query selection quality can be gradually enhanced for efficient planning. Extensive ablations showcase the superiority of our proposed \textbf{Ego-Centric Fully Sparse Paradigm}.

\section{Limitations and Social Impact}
\label{app:limit_impact}

\noindent
\textbf{Limitations.} There still remains some limitations in our work. First, EgoFSD performs ego-centric query selection based on the geometric score, which is obtained through referring to the distance map using static query positions. And the motion information of surrounding agents are introduced into the ego-object dual interaction layer \textbf{implicitly} through iteratively updating the agent queries. How to generate better geometric score considering agent dynamics \textbf{explicitly} is worthy of further discussion. Besides, how to incorporate additional traffic signals and vision-language models into current driving system for autonomous decision making also deserves further exploration.

\noindent
\textbf{Social Impact.} EgoFSD could be easily deployed on mass-produced car chips with different limits of computing resources, and thus can be served as a plug-and-play software to assist human drivers in decision-making and safe driving.


\section{Qualitative Results}

We visualize the motion trajectories of sparse interactive agents as well as planning results of EgoFSD as illustrated in Fig.~\ref{fig_visualization}. Both surrounding camera images and prediction results on BEV are provided accordingly. Besides, we also project the planning trajectories to the front-view camera image. Only the top-3 trajectories of selected agents interacting with ego-vehicle are visualized for better understanding of EgoFSD motivation. EgoFSD outputs planning results based on the vectorized representation in an end-to-end manner, not requiring any dense interaction and redundant motion modeling, let alone hand-crafted post-processing.

\begin{figure*}
	\centering
	\includegraphics[width=1.0\linewidth]{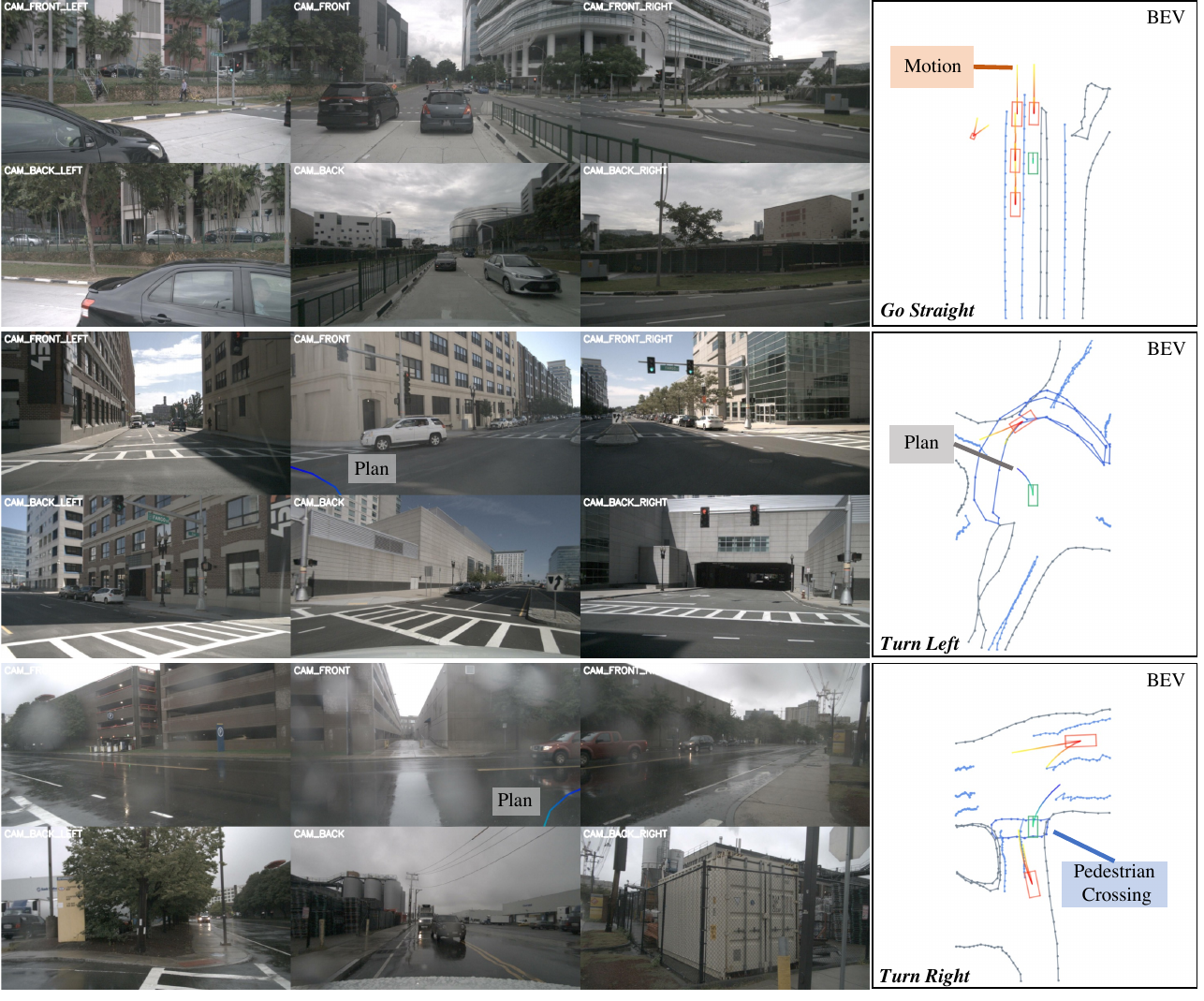}
 \vspace{-5pt}
	\caption{Qualitative results of EgoFSD. EgoFSD outputs planning results based on hierarchical interaction and joint motion of sparse interactive agents without considering other irrelevant objects. We omit the map selection results for clarity of road structure details.}
	\label{fig_visualization}
\vspace{-5pt}
\end{figure*}

\begin{figure*}
\vspace{-5pt}
	\centering
	\includegraphics[width=1.\linewidth]{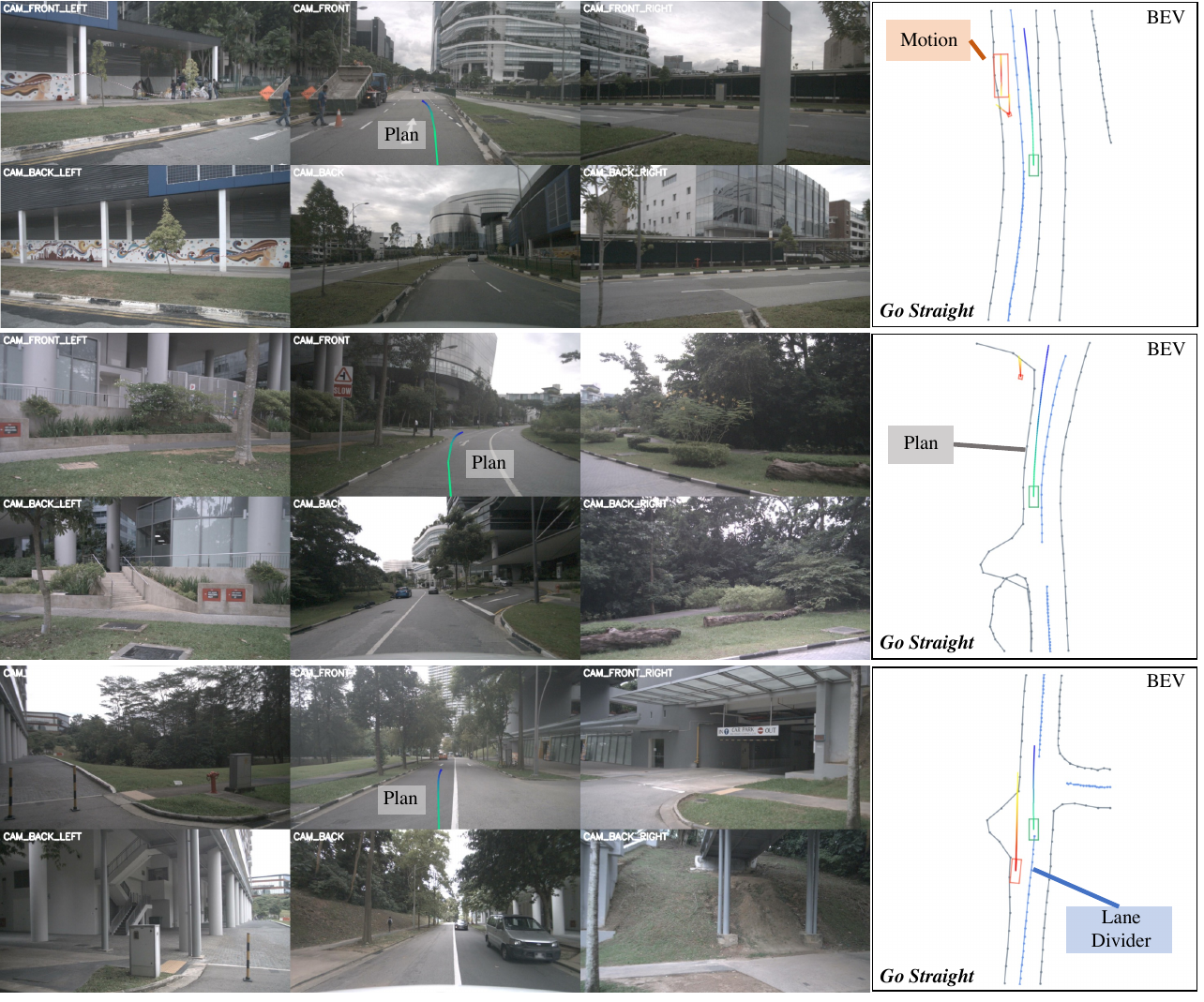}
	\caption{Qualitative results of EgoFSD under ``\textit{Go Straight}" driving command in interactive scenes. In the first row, the pedestrian and the  construction vehicle are selected as the closest in-path agents for motion prediction and interactive planning, thus EgoFSD adjusts the planned trajectory from afar to avoid a collision. In the second row, EgoFSD notices the pedestrian in the distance and plans the future trajectory taking the pedestrian intention into consideration. In the third row, EgoFSD completes interactive decision-making in the ``Cut-in" scenario, and outputs the planned trajectory constrained by the lane divider.} %
	\label{fig_vis_forward}
\vspace{-5pt}
\end{figure*}

\begin{figure*}
\vspace{-5pt}
	\centering
	\includegraphics[width=1.\linewidth]{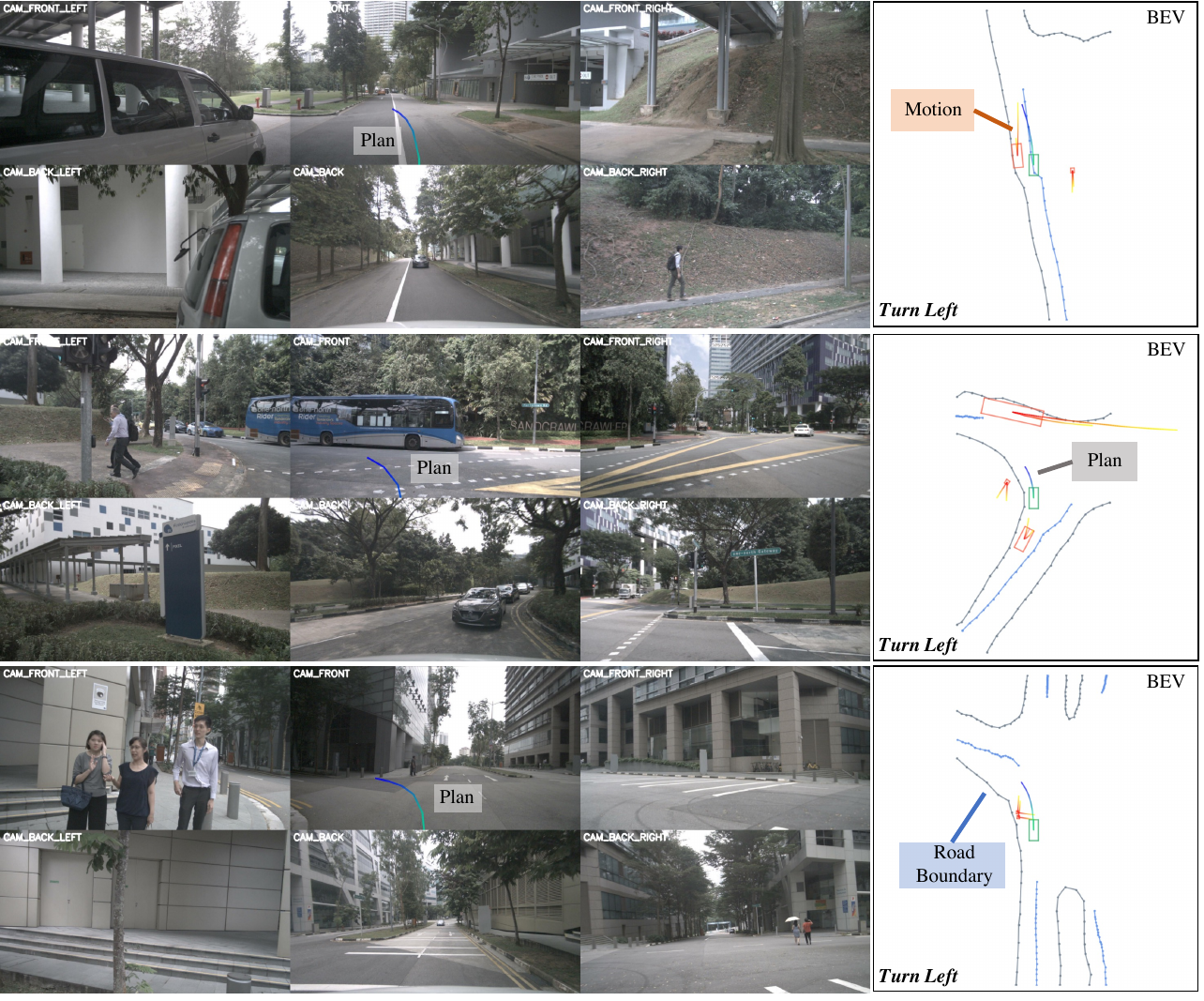}
	\caption{Qualitative results of EgoFSD under ``\textit{Turn Left}" driving command in diverse scenarios. In the first scenario, EgoFSD makes an ``Overtaking" decision from the ride side of the front vehicle, considering the motions of both target vehicle and neighboring pedestrian to ensure driving safety. In the latter two intersection scenarios, EgoFSD detects the pedestrians waiting at the crossing and the opposite bus passing the intersection, then decelerates to make a turning decision.} %
	\label{fig_vis_left}
\vspace{-5pt}
\end{figure*}

\begin{figure*}
\vspace{-5pt}
	\centering
	\includegraphics[width=1.\linewidth]{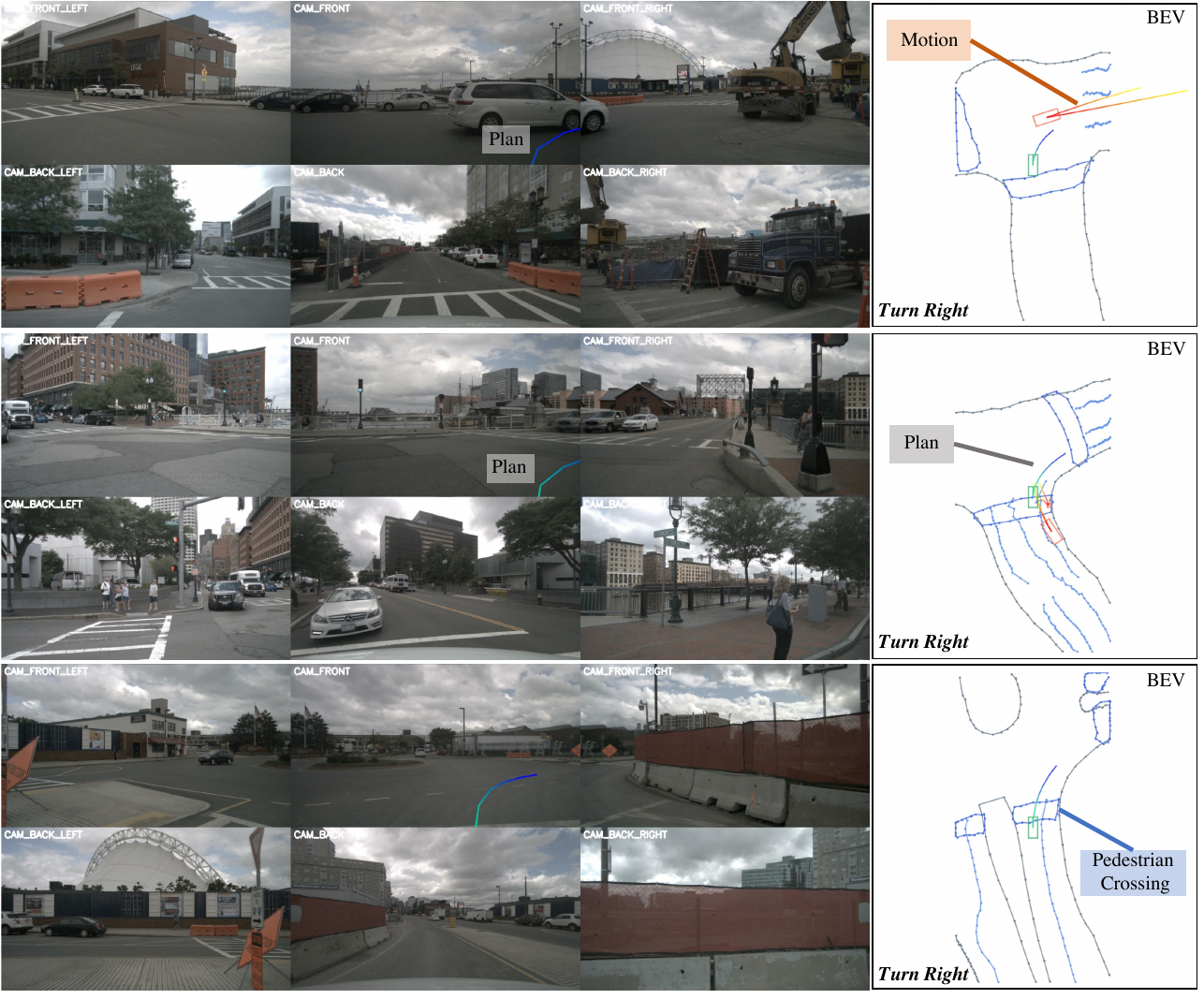}
	\caption{Qualitative results of EgoFSD under ``\textit{Turn Right}" driving command at both interactive and non-interactive intersections. Joint motion prediction of agents and ego-vehicle is essential for EgoFSD especially in the turning scenarios at interactions. The first two rows illustrate the interactive scenarios either inside and outside the intersection. And the last row presents a non-interactive intersection where EgoFSD plans the future trajectory merely based on the detected pedestrian crossing.} %
	\label{fig_vis_right}
\vspace{-5pt}
\end{figure*}

\begin{figure*}
\vspace{-5pt}
	\centering
	\includegraphics[width=1.\linewidth]{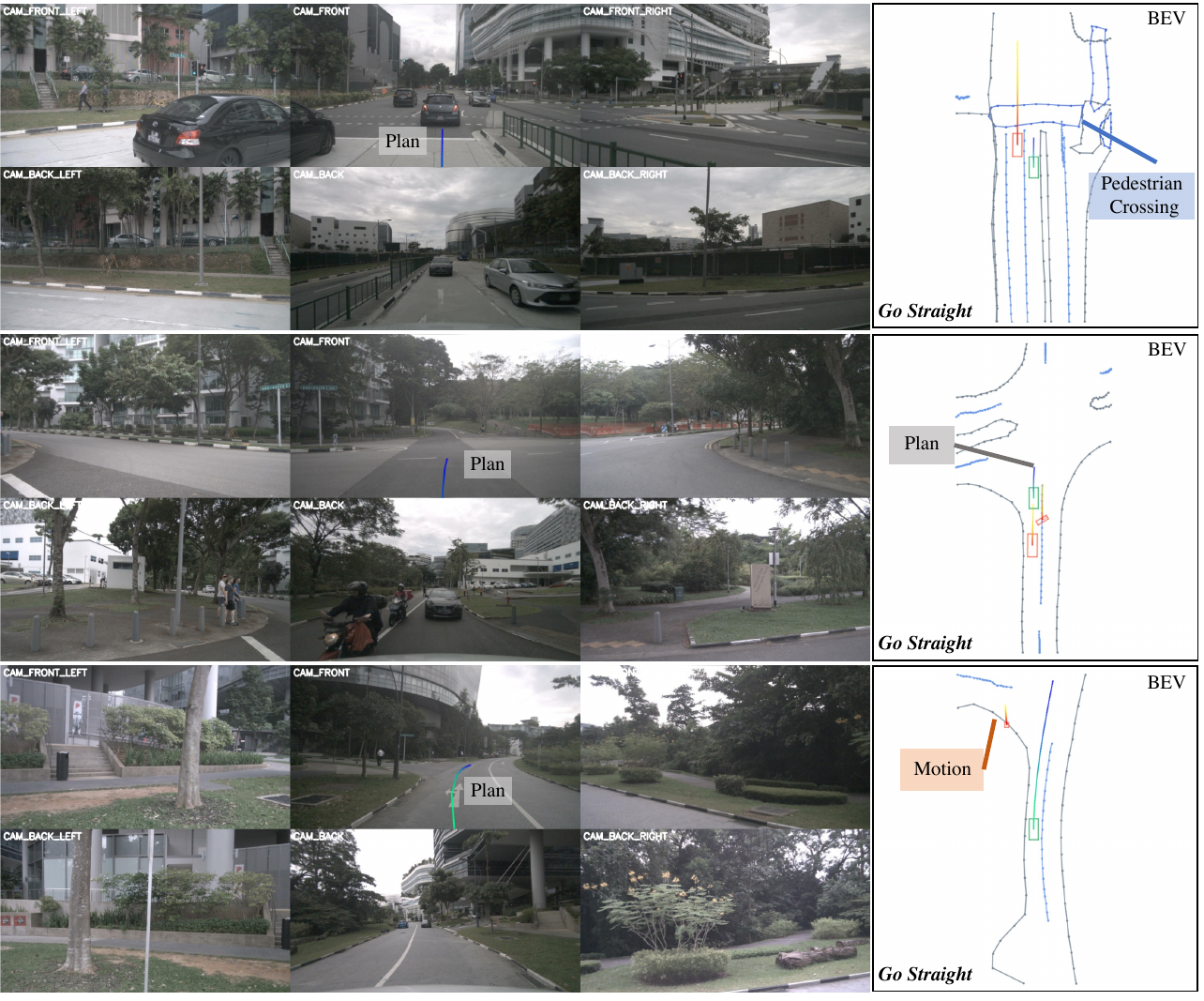}
	\caption{Failure cases of EgoFSD with stationary status. The ego-vehicle is found to remain stationary in either crowded ``car-following" or spacious ``intersection-crossing" scenarios, while EgoFSD still outputs a straight-ahead planning result. However, from the selected perception and motion results, moving forward in these scenarios is also an alternative choice and more reasonable. This also indicates our EgoFSD doesn't depend on the previous ego status during the planning stage, without introducing the motion priors of ego vehicle.} %
	\label{fig_vis_stop}
\vspace{-5pt}
\end{figure*}

\begin{figure*}
\vspace{-5pt}
	\centering
	\includegraphics[width=1.\linewidth]{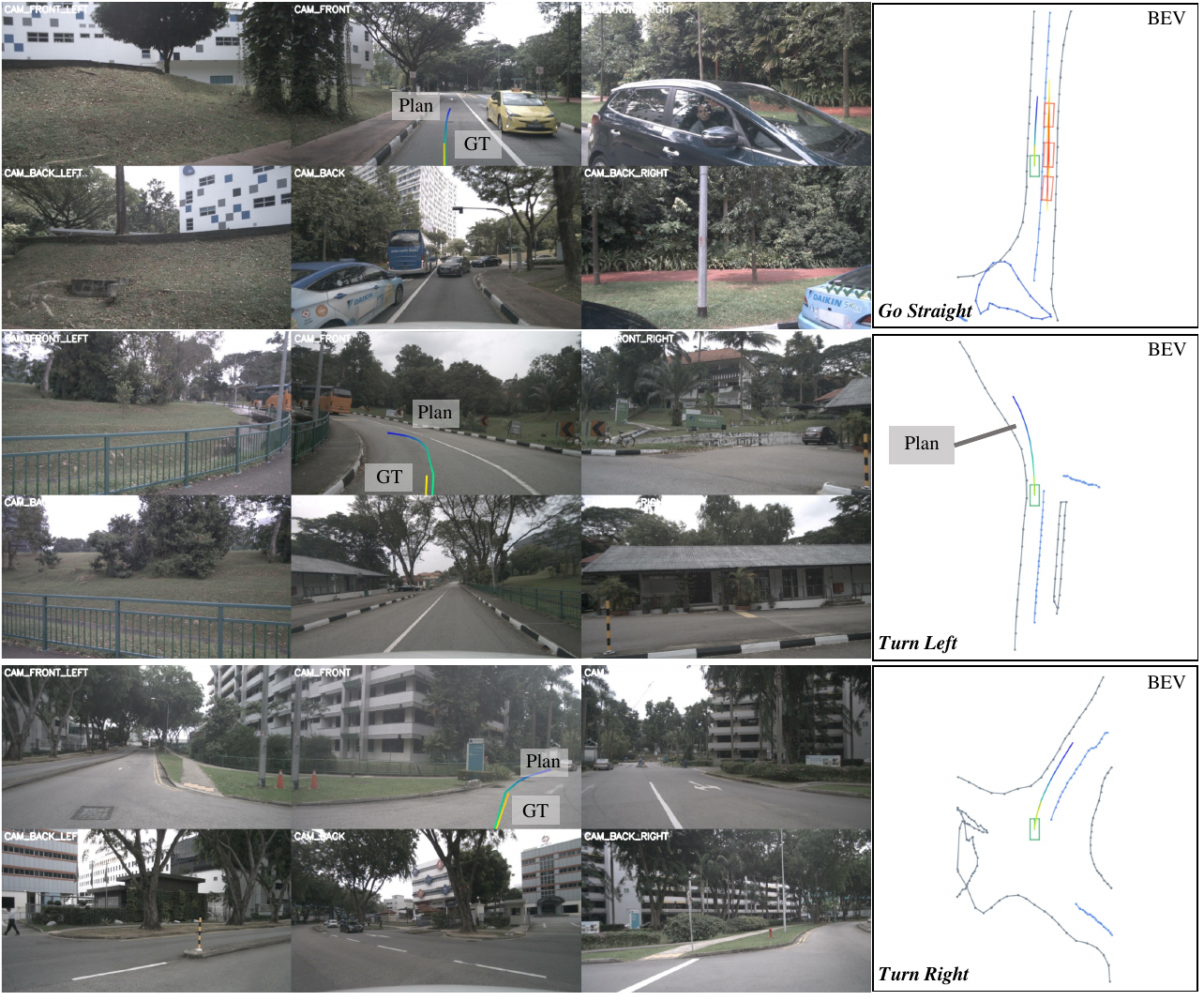}
	\caption{Failure cases of EgoFSD with large average L2 error. As can be seen in various scenarios, our EgoFSD can make a timely response to different driving commands and plan a more efficient future trajectory of ego vehicle compared with the recorded one of expert driver, considering the driving safety, efficiency and comfortness simultaneously, with the help of the proposed ego-centric fully sparse paradigm.} %
	\label{fig_vis_badcase}
\vspace{-5pt}
\end{figure*}

As show in Fig.~\ref{fig_vis_forward}-\ref{fig_vis_right}, we provide more visualization results to illustrate the effectiveness of EgoFSD on various driving scenarios (\textit{i.e.}, interactive / non-interactive scenes, lane-change / lane-keep / following scenes, overtaking / avoidance scenes, intersection scenes) under different commands (\textit{i.e.}, ``Go Straight", ``Turn Left", ``Turn Right"). Moreover, we also observe some failure cases on nuScenes validation set as illustrated in Fig.~\ref{fig_vis_stop} and Fig.~\ref{fig_vis_badcase}. The corresponding explanations as well as analysis are described in the captions below respectively, which mainly attribute to the irrationality of the expert (ground truth) future trajectories of ego vehicle, demonstrating the strong generalizability of our proposed EgoFSD in planning the efficient and reasonable results based on the selected sparse representations of surrounding driving scenarios.

\end{document}